%% file: main.tex
\documentclass[UTF8]{article} 
\usepackage{iclr2023_conference,times}

\usepackage{microtype}
\usepackage{amsmath}
\usepackage{amssymb}
\usepackage{booktabs}
\usepackage{colortbl}
\usepackage{CJKutf8}
\usepackage[utf8]{inputenc}
\definecolor{lightgray}{rgb}{0.9,0.9,0.9}
\usepackage{caption}
\usepackage{multicol}
\usepackage{wrapfig}
\usepackage{subcaption}
\usepackage{xcolor}
\usepackage{graphicx}
\usepackage{setspace}
\usepackage{hyperref}
\usepackage{url}
\usepackage{multirow}
\usepackage{colortbl}
\usepackage{tabularx}
\usepackage{blindtext}
\usepackage{pgfplots}
\pgfplotsset{compat=1.18} 
\usepackage{tikz}
\usetikzlibrary{er,positioning,bayesnet}
\usepackage[inline]{enumitem}
\usepackage{makecell}
\usepackage{tipa}
\usepackage{siunitx}
\usepackage{tocloft}
\usepackage{listings}
\usepackage[raster,skins]{tcolorbox} 
\usepackage{xltabular}
\usepackage{hyperref}
\usepackage[framemethod=tikz]{mdframed}
\surroundwithmdframed[
  hidealllines=true,
  innerleftmargin=0pt,
  innertopmargin=0pt,
  innerbottommargin=0pt]{lstlisting}
\lstnewenvironment{response}[1][] 
 {\lstset{
 columns=fullflexible,
  breakautoindent=false, breakindent=0pt, breaklines, linewidth=8cm, #1}}
 {}

\newcommand{\method}{\textbf{Waver}\xspace}

\newlength\savewidth\newcommand\shline{\noalign{\global\savewidth\arrayrulewidth
  \global\arrayrulewidth 1pt}\hline\noalign{\global\arrayrulewidth\savewidth}}

\definecolor{url_color}{RGB}{113, 187, 179}

\hypersetup{
    colorlinks=true,
    linkcolor=black,
    urlcolor=url_color,
}

\title{Waver: Wave Your Way to Lifelike Video Generation}
\author{
\textbf{Bytedance Waver Team}
}

\iclrfinalcopy
\begin{document}

\maketitle

\begin{abstract}

We present \method, a high-performance foundation model for unified image and video generation. \method can directly generate videos with durations ranging from 5 to 10 seconds at a native resolution of 720p, which are subsequently upscaled to 1080p. The model simultaneously supports text-to-video (T2V), image-to-video (I2V), and text-to-image (T2I) generation within a single, integrated framework. We introduce a Hybrid Stream DiT architecture to enhance modality alignment and accelerate training convergence. To ensure training data quality, we establish a comprehensive data curation pipeline and manually annotate and train an MLLM-based video quality model to filter for the highest-quality samples. Furthermore, we provide detailed training and inference recipes to facilitate the generation of high-quality videos.
Building on these contributions, \method excels at capturing complex motion, achieving superior motion amplitude and temporal consistency in video synthesis. Notably, it ranks among the Top 3 on both the T2V and I2V leaderboards at Artificial Analysis (data as of 2025-07-30 10:00 GMT+8), consistently outperforming existing open-source models and matching or surpassing state-of-the-art commercial solutions. We hope this technical report will help the community more efficiently train high-quality video generation models and accelerate progress in video generation technologies. Official page: \url{https://github.com/FoundationVision/Waver}.

\end{abstract}


\begin{figure}[htbp]
    \centering
    \includegraphics[width=0.9\textwidth]{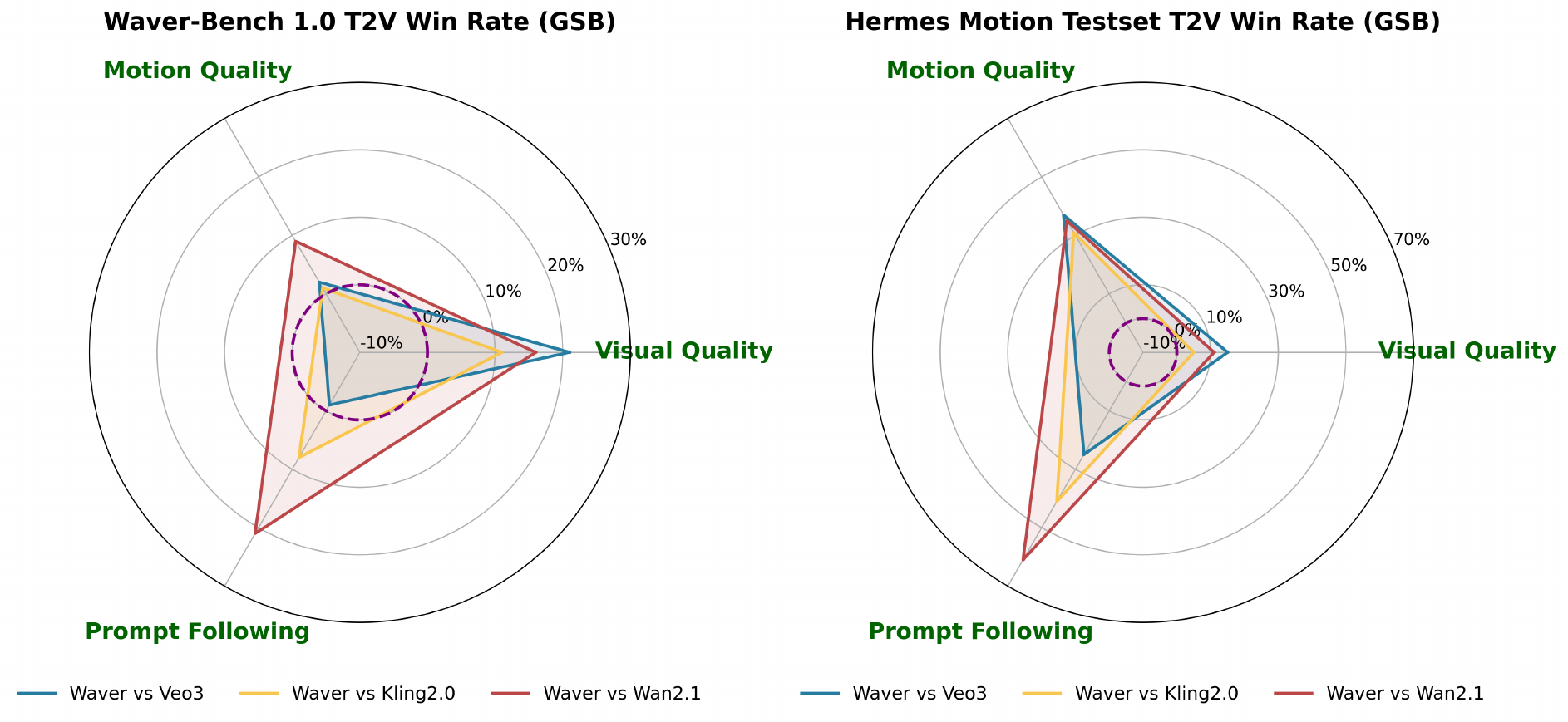}
    \caption{
        \textbf{Left: }Human evaluation win rates (GSB) of \method compared to Veo3, Kling2.0, and Wan2.1 on the Waver-bench 1.0 Text-to-Video (T2V) dataset across three dimensions: motion quality, visual quality, and prompt following. Waver-bench 1.0 covers a wide range of scenarios, including sports, daily activities, landscapes, animals, animations, and more. \textbf{Right: }Human evaluation win rates (GSB) on the Hermes Motion Testset across three dimensions: motion quality, visual quality, and prompt following. The Hermes Motion Testset encompasses complex motion scenarios such as tennis, basketball, gymnastics, and others. We can observe that, compared to general scenarios, \method demonstrates a more pronounced advantage in complex motion scenarios.
    }
    \label{fig:common_eval}
\end{figure}

\clearpage

\newpage

\newcommand{\customsize}{\fontsize{7.7}{9}\selectfont}

\begingroup
\footnotesize
\begin{spacing}{0.88}
\tableofcontents
\end{spacing}
\endgroup

\hypersetup{linkcolor=url_color}

\newpage

\begin{figure}[htbp]
    \centering
    \includegraphics[width=\textwidth]{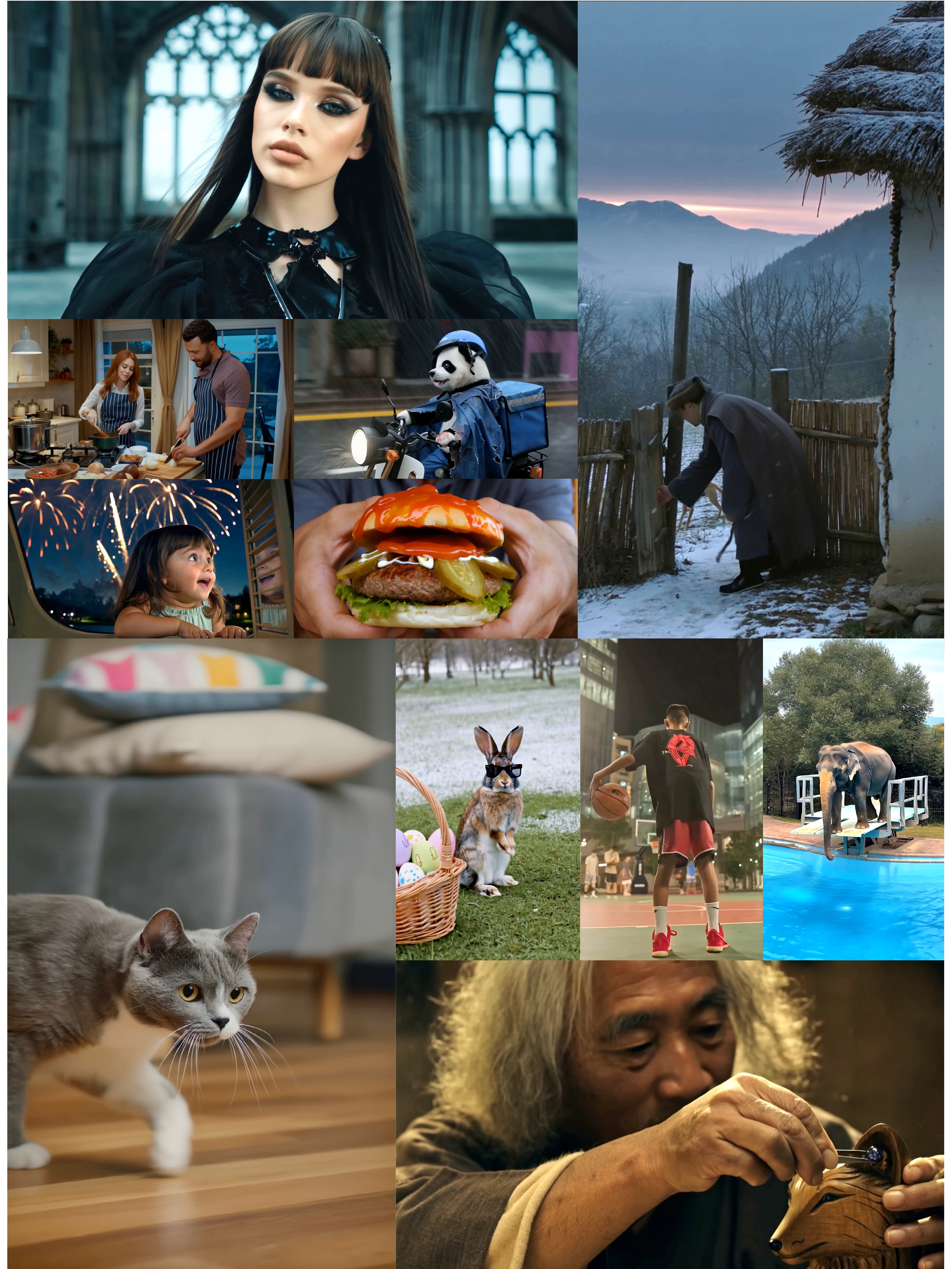}
    \caption{
        T2V samples generated by \method. \method is capable of generating 1080p videos at arbitrary aspect ratios, delivering high levels of aesthetic quality, realism, and motion fidelity, while simultaneously supporting both T2V and I2V tasks.
    }
    \label{fig:demo}
\end{figure}

\clearpage

\newpage


\input{content/1_intro}

\input{content/2_model_arch}

\input{content/3_data}

\input{content/4_recipe}

\input{content/5_infra}

\input{content/6_bench}

\input{content/7_discussion}

\input{content/8_conclusion}

\input{content/9_authors}

\bibliography{main}
\bibliographystyle{iclr2023_conference}
\clearpage

\end{document}

%% file: content/1_intro.tex
\section{Introduction}
\label{sec:intro}

Video generation~\citep{singer2022make,blattmann2023stable,openaisora2024} is currently a highly active area of interest in both academia and industry. It not only fulfills the limitless creative aspirations people have for video content but is also gradually being applied to a range of commercial scenarios, such as digital human live streaming for e-commerce, product display, and virtual try-on applications. Existing video generation models, most of which are based on the DiT~\citep{dit} architecture, have achieved remarkable progress in a short period of time. Open-source models such as Wan~\citep{wan2025wan}, StepVideo~\citep{ma2025step}, HunyuanVideo~\citep{kong2024hunyuanvideo}, and CogVideoX~\citep{cogvideox} have greatly advanced the development of the video generation community. Proprietary models such as Veo~\citep{veo32025}, Kling~\citep{kuaishou2024kling}, Hailuo~\citep{minimax2024hailuo}, and Seedance~\citep{gao2025seedance} demonstrate better performance. 

However, the current video generation community still faces numerous challenges: (1) The quality of generated videos remains unsatisfactory, often lacking aesthetic appeal and realism, with poor performance in complex motion scenarios such as gymnastics and basketball; (2) Existing public technical reports rarely discuss the technical details required for generating high-resolution videos, such as those at 1080p, and the rationale behind certain model architecture designs is often unclear; (3) Most video generation models~\citep{kong2024hunyuanvideo,wan2025wan} implement text-to-video (T2V) and image-to-video (I2V) tasks using two separate models, resulting in substantial training resource overhead; (4) There is limited information regarding data processing pipelines and model training procedures, making it difficult to fully reproduce the entire training process.

To overcome these problems and challenges, we propose \method, a model based on Rectified Flow Transformer~\citep{esser2024sd3} and meticulously designed to achieve industry-grade performance. \method consists of two modules: Task-Unified DiT and Cascade Refiner. We first employ Task-Unified DiT to generate videos with a resolution of 720p and durations ranging from 5 to 10 seconds, and subsequently apply Cascade Refiner to upscale them to 1080p resolution.

\textbf{Task-Unified DiT: Joint modeling of multiple tasks allows effective learning of task-specific features and mutual promotion. Hybrid Stream enhances modality alignment and accelerates convergence.}
In Task-Unified DiT, we employ a flexible input conditioning mechanism to unify text-to-image (T2I), text-to-video (T2V), and image-to-video (I2V) tasks within a single network. It only needs to modify the channel dimension of the model input. We further design a Hybrid Stream DiT architecture to optimize the trade-off between modality alignment and parameter efficiency, while also accelerating training convergence. At this stage, we have achieved high-quality video generation at a native resolution of 720p, alongside image generation at resolutions up to 1080p. 

\textbf{Cascade Refiner: With reduced computational overhead, videos are upscaled to 1080p, significantly enhancing visual clarity.}
In the second stage, the Cascade Refiner takes the 720p videos generated in the first stage as input and outputs 1080p videos with identical content but significantly enhanced clarity. This process is also implemented via flow matching with fewer inference steps, enabling the model to learn the transition from low-resolution to high-resolution videos. Compared to single-stage methods that directly generate 1080p videos, our two-stage approach achieves a 40\% acceleration.

\textbf{Comprehensive data pipelines and detailed training recipes reveal the full details of video generation model training.}
We establish a comprehensive data curation pipeline, which includes manually annotated and trained caption and quality models to obtain high-quality training data and captions. In total, our model has seen more than 200 million video clips. We also provide detailed recipes outlining how to optimize the model's semantic representation, aesthetics, motion, and realism, as well as how to balance these different aspects. Additionally, we present the data volume used at each training stage, multiple hyperparameters and noise scheduling for both training and inference, and infrastructure optimization strategies.

\textbf{Superior generation quality in both general scenes and complex motion scenarios.}
We conduct a comprehensive evaluation of \method across multiple dimensions and benchmarks. On the public leaderboard of the Artificial Analysis Arena, \method ranks third in both the T2V and I2V tracks, with data as of 2025-07-30 10:00 (GMT+8). On our internal Waver-bench 1.0, \method outperformed Kling2.0 and Wan2.1~\citep{wan2025wan} in terms of motion quality, visual quality, and prompt following; it also surpasses Veo3 in motion quality and visual quality, while its prompt following was slightly weaker than Veo3. On our challenging Hermes Motion Testset, specifically designed for complex motion evaluation, \method demonstrates significant improvements over Veo3, Kling2.0, and Wan2.1 in both motion quality and prompt following. We hope that the optimization details presented in this technical report can help the community further advance the performance of current video generation models.

%% file: content/2_model_arch.tex
\section{Model Architecture}
\label{sec:model-arch}

\begin{figure}[htbp]
    \centering
    \includegraphics[width=1.0\textwidth]{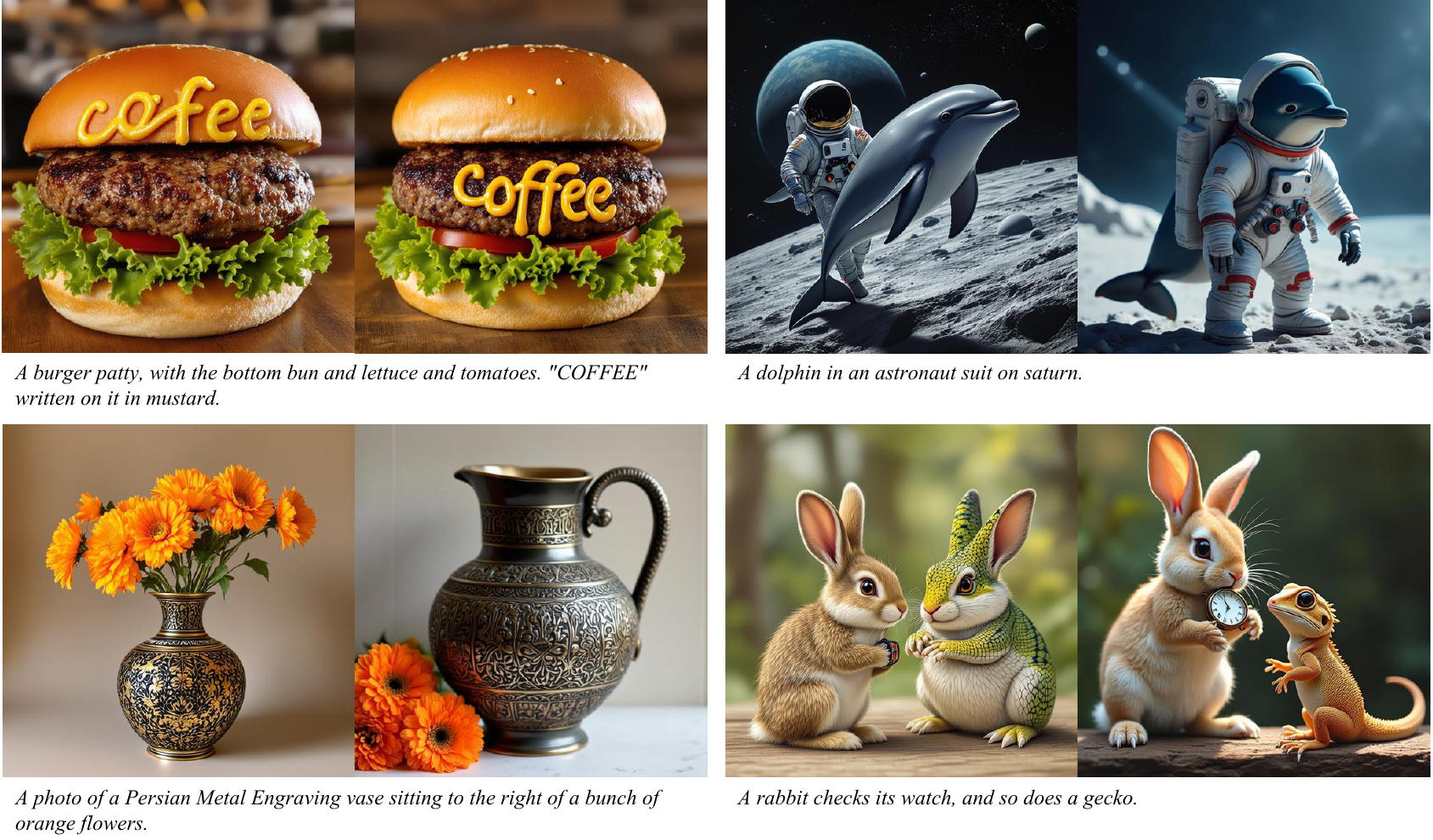}
    \caption{
        Comparison of 512p T2I results using different text encoders. For each case, image on the \textbf{left} is only using flan-t5-xxl and that on the \textbf{right} is using flan-t5-xxl and Qwen2.5-32B-Instruct. 
    }
    \label{fig:t2i_qwen}
\end{figure}

We use Wan2.1-VAE~\citep{wan2025wan} for efficient video latent compression and a powerful dual-encoder system, combining flan-t5-xxl~\citep{t5} and Qwen2.5-32B-Instruct~\citep{qwen2.5}, for superior text understanding. 
Benefiting from the strong performance of Qwen, this dual-encoder setup demonstrably improves prompt following in the text-to-image (T2I) task over the single-encoder (flan-t5-xxl) baseline (Fig.~\ref{fig:t2i_qwen}). 
The architecture of \method primarily consists of two components: the Task-Unified DiT and the Cascade Refiner. 
The Task-Unified DiT is built upon rectified flow Transformers~\citep{esser2024sd3}, serves as the core generation model. 
It fuses video and text modalities through a "Dual Stream + Single Stream" architecture.
The Cascade Refiner is also based on rectified flow Transformers, functions as a super-resolution module. 
It takes the 480p or 720p videos generated by the Task-Unified DiT as input and upscales them to a final resolution of 1080p. 
This design enables efficient, high-fidelity video super-resolution with fewer inference steps.

\begin{figure}[htbp]
    \centering
    \includegraphics[width=1.0\textwidth]{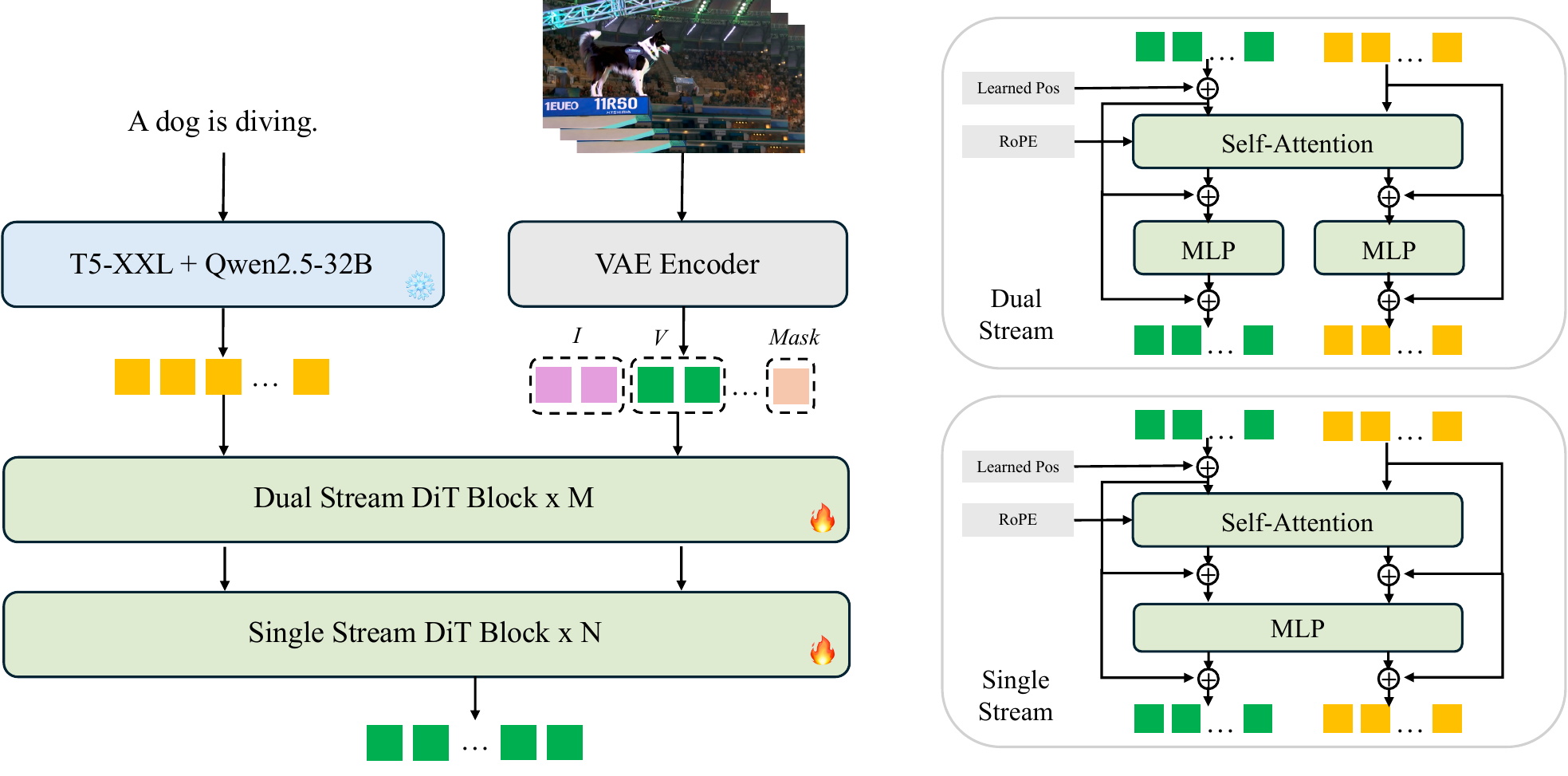}
    \caption{
        Architecture of Task-Unified DiT.
    }
    \label{fig:dit}
\end{figure}

\subsection{Task-Unified DiT}

\paragraph{Unified Input Formulation.}
To unify diverse generative tasks (T2I, T2V, I2V) within a single framework, we employ a flexible input conditioning mechanism based on a three-part input tensor. This input is formed by concatenating a primary noisy latent ($V$), a conditional frames tensor ($I$), and a binary condition mask ($Mask$) along the channel dimension.
The tensor $I$ contains the VAE-encoded latents of any known conditioning frames, with other frames filled by black image latents. The $Mask$ tensor then specifies which frames are conditions (value 1) and which are to be generated (value 0), providing a unique binary indicator for each frame in the sequence.
This formulation is highly flexible. It not only allows us to mix tasks and adjust their proportions during training, but also enables straightforward extension to other tasks, such as video interpolation.

\begin{figure}[!t]
    \centering
    \includegraphics[width=0.9\textwidth]{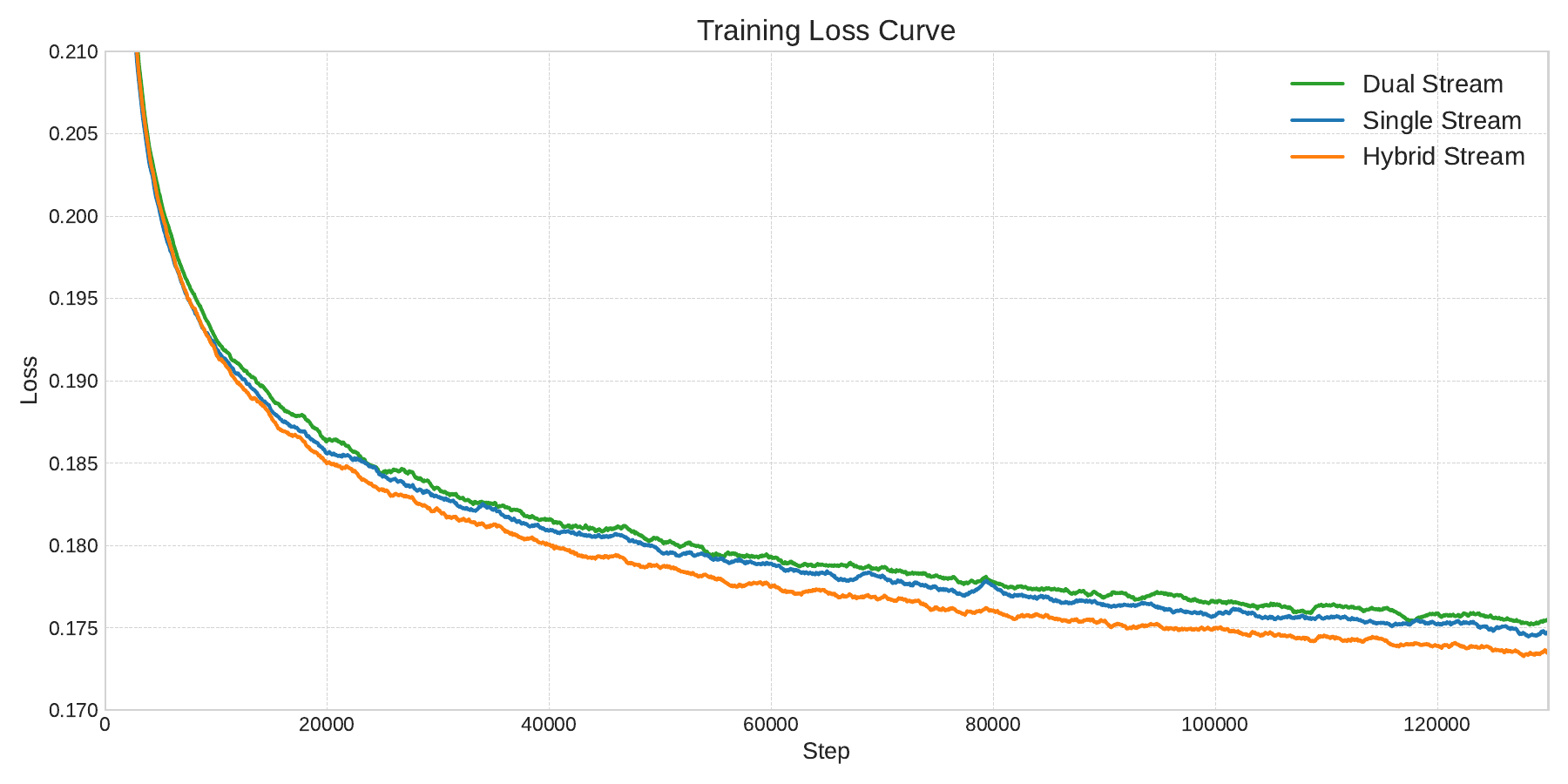}
    \caption{
        Loss comparison between Hybrid Stream, Dual Stream, and Single Stream structures. Hybrid Stream's loss converges faster.
    }
    \label{fig:hybrid-loss}
\end{figure}

\paragraph{Hybrid Stream Structure. }
Our Task-Unified DiT combines Dual Stream and Single Stream blocks to optimize the trade-off between modality alignment and parameter efficiency.
The Dual Stream block processes video and text modalities with separate parameters, merging them only during self-attention. 
This allows for co-adaptation of text and video features, fostering strong alignment, but at the cost of parameter efficiency; in text-to-video generation, the large number of video tokens means the text-specific parameters are underutilized.
Conversely, the Single Stream block uses a shared set of parameters for the combined modalities, maximizing efficiency. 
However, this shared approach can slow convergence due to the differing statistical distributions of video and text data.
Therefore, we adopt a hybrid strategy: the first M layers use the Dual Stream design to effectively align the modalities, while the subsequent N layers switch to the Single Stream design for computational efficiency.
This Hybrid Stream structure demonstrably achieves faster convergence than either pure approach, as evidenced in Fig.\ref{fig:hybrid-loss}.
Some key parameters can be found in Tab.~\ref{tab:dit_parameters}.


\begin{table}[!t]
\centering
\begin{tabular}{lcccccc}
\toprule 
\textbf{Model Size} & \textbf{M} & \textbf{N} & \textbf{Input Dim.} & \textbf{Output Dim.} & \textbf{Num of Heads} & \textbf{Head Dim} \\
\midrule 
12B & 16 & 40 & 36 & 16 & 24 & 128 \\
\bottomrule 
\end{tabular}
\caption{Key parameter selections in our Task-Unified DiT.}
\label{tab:dit_parameters}
\end{table}

\paragraph{Hybrid Position Embedding. }
To effectively encode spatio-temporal information, we employ a hybrid positional encoding scheme that synergizes relative and absolute position signals. 
This scheme combines 3D Rotary Position Embedding (RoPE)~\citep{su2024roformer} with a factorized learnable position embedding~\citep{zhai2022scaling}. 
The first component, 3D RoPE, provides relative positional information across three dimensions: temporal, height, and width.
In contrast to the sinusoidal encoding used in ViT~\citep{dosovitskiy2020image}, 3D RoPE offers stronger extrapolation capability for varying video durations and resolutions and reduces distortion and deformation in generated videos. 
The second component, the factorized learnable positional embedding, supplies absolute position information.
It independently encodes each dimension (temporal, height, width) and then sums the resulting embeddings. 
By providing an explicit positional anchor, this method has been shown to accelerate model convergence during training.

\subsection{Cascade Refiner}
\label{sec:refiner}

Training and inference on 1080p videos are computationally expensive. 
To address this, we propose a two-stage approach inspired by~\citep{zhang2025flashvideo}. 
In the first stage, we generate a low-resolution video. 
In the second stage, this video is upscaled to 1080p using a dedicated refiner model. 
This hierarchical process can potentially accelerate the generation of 1080p videos, as the refiner is conditioned on the strong priors provided by the initial low-resolution video.

\begin{figure}[htbp]
    \centering
    \includegraphics[width=1.0\textwidth]{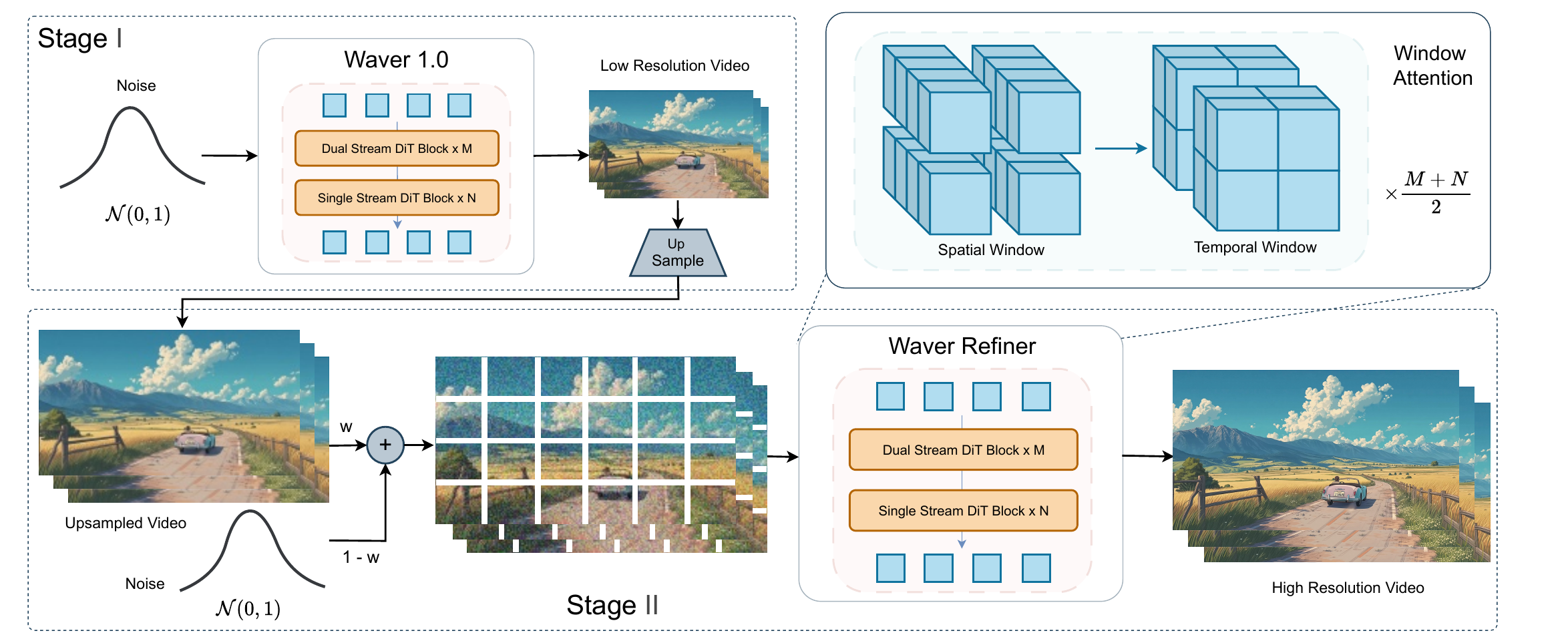}
    \caption{
        Pipeline of Cascade Refiner.
    }
    \label{fig:cascade}
\end{figure}

To this end, we trained a video refiner that takes a 480p or 720p video generated in the first stage as input and outputs a 1080p video with identical content but significantly enhanced clarity. 
Architecturally, the refiner is based on Waver1.0, but we have replaced the standard attention mechanism with window attention~\citep{seawead2025seaweed}.
For training this refiner, we constructed low-resolution and high-resolution video pairs by applying a degradation process to our high-definition training data. 
We then employ a flow matching technique to learn the transition from the low-resolution video distribution to the high-resolution video distribution. 
At inference time, the output video from the first stage undergoes this same degradation process. This degraded video is then fed into the refiner to generate the final, polished 1080p high-definition result.
A complete overview of this two-stage pipeline is depicted in Fig. \ref{fig:cascade}.

\paragraph{Window Attention}
We improve the efficiency of 1080p video generation by leveraging window attention to reduce computational costs.
This mechanism partitions video tokens into local $W_h \times W_w \times W_t$ windows and restricts attention calculations to within these boundaries.
To enable cross-window communication, we alternate between spatial (2x2x1 window) and temporal (1x1x2 window) attention schemes.
Furthermore, we observed that attention patterns in the DiT architecture are densest in the shallowest and deepest layers.
We therefore adopt a hybrid approach, applying full attention to the first and last eight layers while retaining window attention for the remaining middle layers.
This method effectively balances high-fidelity video synthesis with computational tractability.

\paragraph{Pixel and Latent Degradation}
To train our super-resolution model, we synthesize low-quality videos from high-quality sources using a two-part degradation process: pixel degradation and latent degradation.
First, for pixel degradation, we downsample 1080p videos to 360p and then upscale them back to 1080p. 
This process simulates the low-resolution output of our first-stage generator.
Second, first-stage outputs often contain generative artifacts and distortions not found in simple downsampled videos. 
To bridge this domain gap, we perform latent degradation by injecting noise into the VAE latent space. 
The noisy latent, $\mathbf{x}_\textbf{n}$, is computed as:
$$\mathbf{x}_n = (1 - w_d) \cdot \mathbf{x} + w_d \cdot \mathbf{n}.$$
Here, $\mathbf{x}$ is the VAE latent of the video with pixel degradation, $\mathbf{n}$ is Gaussian noise, and $w_d$ is a weight sampled from a predefined range in each training step. 
In our experiments, we found that sampling $w_d$ from the range $[0.85, 0.95]$ yielded good performance.
This strategy forces the model to learn robustness against both low resolution and generative distortions, reducing the train-inference distribution mismatch.

\begin{figure}[!t]
    \centering
    \includegraphics[width=1.0\textwidth]{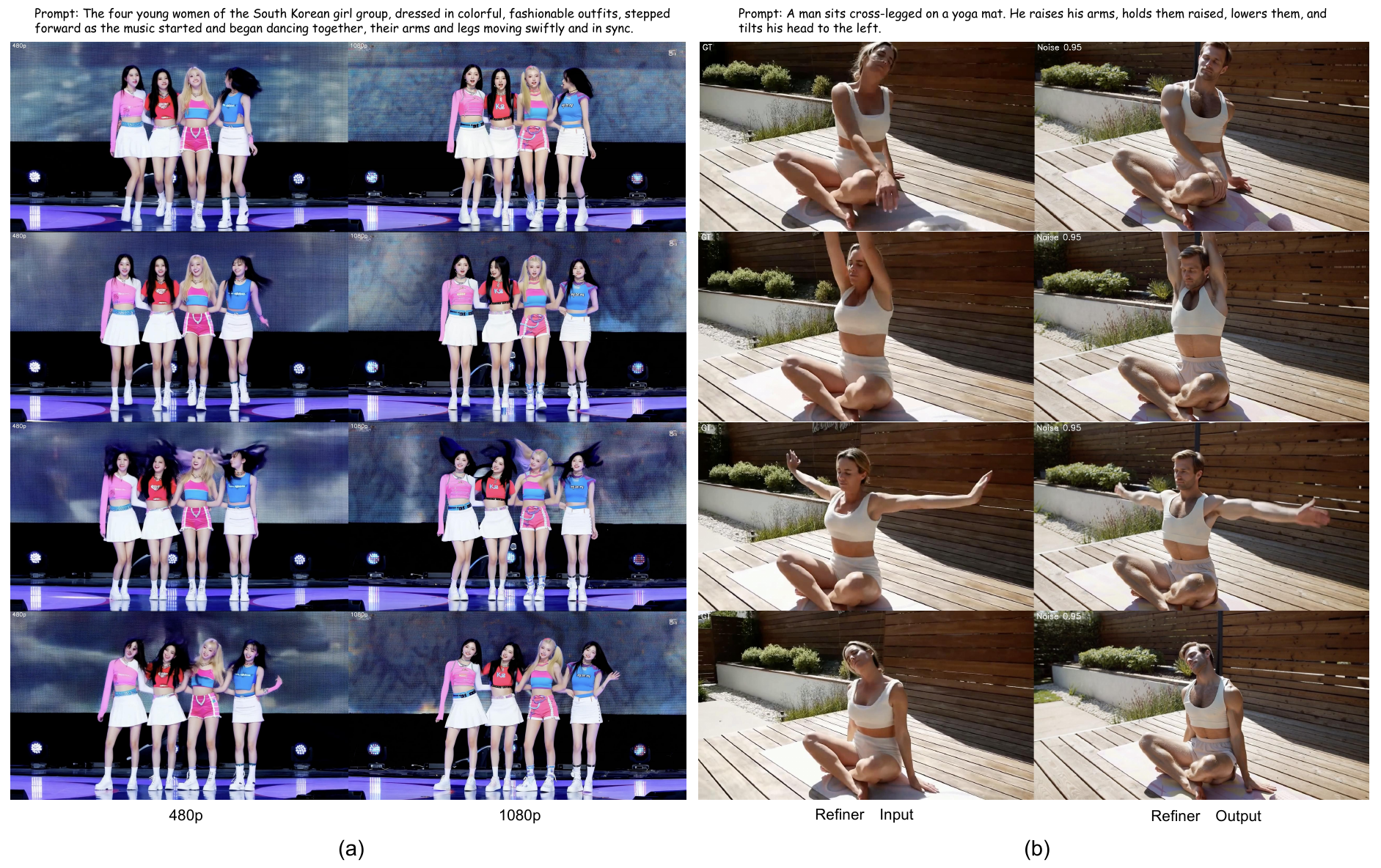}
    \caption{
        Fig.(a) illustrates the Refiner's output, where it has upscaled a 480p video from the first stage to 1080p. Besides improving the sharpness, it has also fixed the visual artifacts. Fig.(b) showcases the Refiner's video editing ability by transforming the woman in the source video into a man. (zoom in for a better view.)
    }
    \label{fig:refiner-example}
\end{figure}

\paragraph{Refiner Training and Inference}
We train the refiner model using a flow matching objective. 
The model input is a linear interpolation between the degraded and clean latents: 
$$\mathbf{x}_{\text{input}} = w \cdot \mathbf{x}_n + (1 - w) \cdot \mathbf{x}_0.$$
The model is trained to predict the degraded latent towards the clean latent: 
$$\mathbf{x}_{\text{target}} = \mathbf{x}_0 - \mathbf{x}_n.$$
Here, $\mathbf{x}_n$ is the degraded VAE latent from our pixel and latent degradation process, $\mathbf{x}_0$ is the ground-truth VAE latent of the high-resolution video, and the weight $w$ is sampled from $[0, 1]$ during training.
During inference, we first generate a low-resolution video using the first stage model. 
This video is then upscaled to 1080p, and its VAE latent undergoes the same noise degradation process described previously. 
This final degraded latent is then fed into the refiner, which outputs the enhanced 1080p video.
Thanks to our comprehensive degradation strategy, the refiner not only enhances video resolution but also corrects generative artifacts and distortions.
Furthermore, we observed that the refiner can perform video editing tasks, such as object modification, when the weight $w_d$ is set to a large value, as illustrated in Fig.\ref{fig:refiner-example}.

%% file: content/3_data.tex
\section{Training Data}

\label{sec:data}
\begin{figure}[htbp] 
    \centering 
    \includegraphics[width=\textwidth]{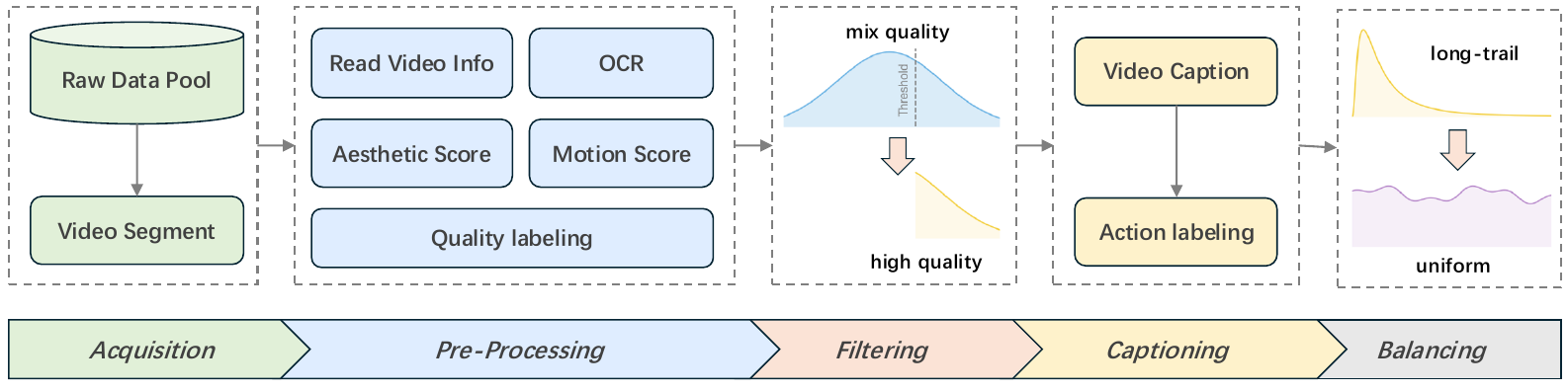} 
    \caption{An overview of our proposed data processing pipeline. The process consists of five main stages: multi-source acquisition, pre-processing, quality filtering, captioning, and semantic balancing.}
    \label{fig:pipeline} 
\end{figure}

The effectiveness of video generation models fundamentally depends on the scale, diversity, and quality of the training data. This process begins with systematic preprocessing and segmentation, which extracts relevant clips from multiple sources to maximize content coverage. This is followed by a hierarchical filtering process that rigorously removes low-quality and unsafe samples. To further enhance data diversity and address potential gaps in real-world content, we also incorporate synthetic data generated through advanced augmentation techniques. This integrated and automated workflow, as illustrated in Figure~\ref{fig:pipeline}, enables us to efficiently construct a robust dataset, laying a solid foundation for training high-performing and generalizable video generation models.

\subsection{Data Pre-Processing}
\label{sec:data_process}
The data preprocessing stage is critical for transforming raw video data into high-quality samples suitable for model training. Through this structured preprocessing workflow, we ensure that only high-quality, diverse, and representative video clips are included in the final dataset, laying a strong foundation for subsequent filtering and augmentation steps. This stage consists of several key steps:

\textbf{Multi-source Video Acquisition.} We collect raw video data from a variety of sources to ensure diversity in content, style, and scenarios. In addition, for video categories that present greater challenges for generative modeling, such as complex ball sports or scenes with intricate motion, we perform targeted data collection and supplementation to address these specific difficulties. This multisource and targeted acquisition strategy helps to mitigate dataset bias and enhances the generalization capacity of the model.

\textbf{Video Segmentation.} The raw videos collected vary in length and are not directly suitable for model training. To address this, we systematically segment them into shorter, manageable clips, following the methodology in ~\citep{chen2025goku}. Specifically, we first leverage PySceneDetect~\citep{pyscenedetect} to perform initial scene detection. We then refine these preliminary segments by extracting DINOv2~\citep{oquab2023dinov2} features from each frame and computing the cosine similarity between adjacent frames. When similarity falls below a set threshold, we mark a shot change and further divide the clip.
Finally, our selection criteria are twofold: (1) we retain all clips with durations between 2 and 10 seconds. (2) For clips exceeding 10 seconds, we subsequently sample segments with the lowest internal DINOv2 feature similarity, as these tend to capture significant motion and highlight moments, ensuring both clip coherence and suitability for training.

\textbf{Clip Quality Scoring.}
Each video clip undergoes a comprehensive evaluation in multiple dimensions to ensure its suitability for model training. This systematic scoring framework allows us to build a high-quality, balanced dataset and supports flexible filtering and selection of clips at different stages of training. The evaluation consists of several key steps:

For basic attributes, we first assess fundamental video properties, such as frame rate (FPS), resolution, and bitrate. Clips that do not meet minimum technical standards are excluded, ensuring that only videos with sufficient clarity and smoothness are retained.

For static visual quality, we use a pretrained aesthetic model to score each frame and employ OCR to measure the proportion of overlaid text, along with watermark detection, to provide quantitative indicators for subsequent filtering.

To assess the dynamic quality of each clip, we analyze its motion magnitude using RAFT~\citep{2021RAFT} for optical flow computation. To disentangle subject motion from camera movement, we first perform foreground segmentation on the clip, and then computed the optical flow separately for the foreground and background regions. The motion magnitude and diversity are subsequently calculated based primarily on the foreground flow, enabling a more precise quantification of subject-specific dynamics.

\subsection{Quality Model}
\label{sec:quality_model}
\begin{figure}[!t]
    \centering 
    \begin{subfigure}{0.48\textwidth} 
        \centering
        \includegraphics[width=\linewidth]{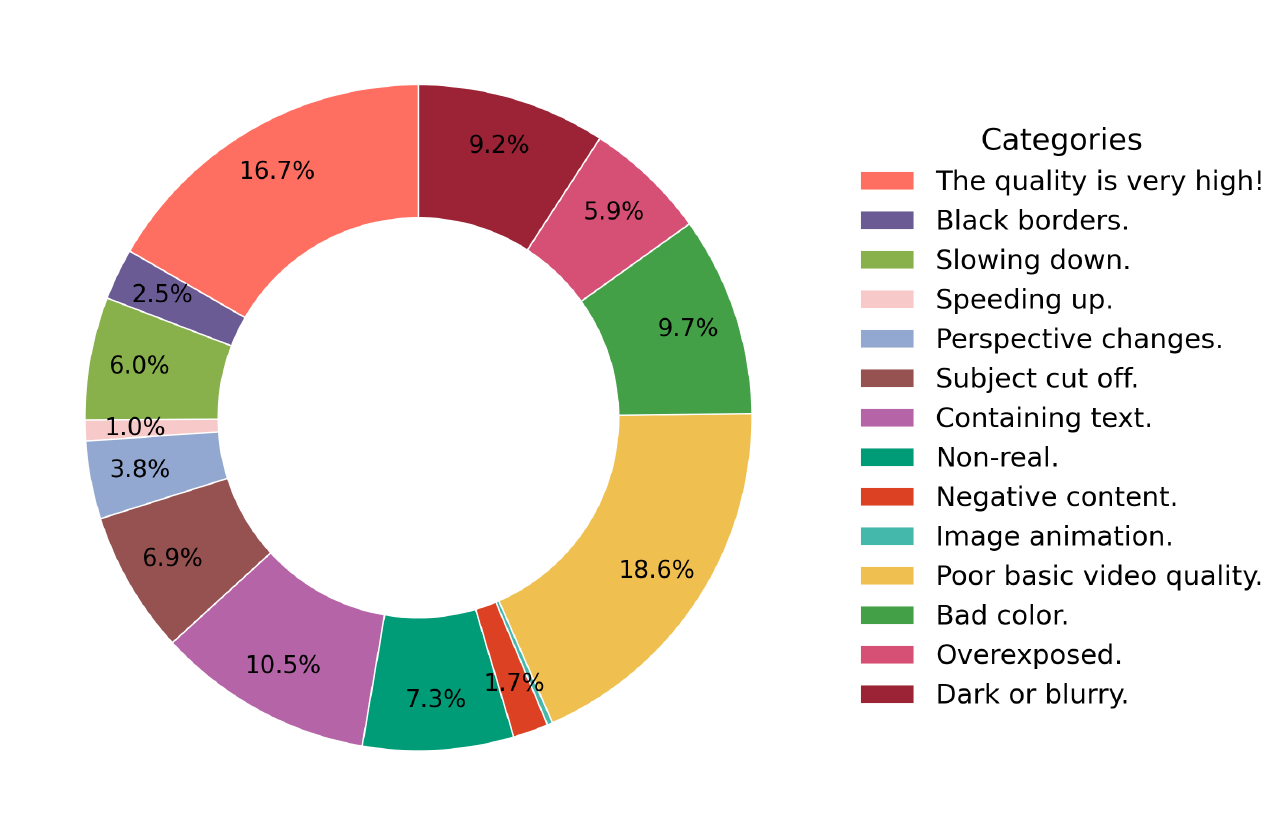}
        \caption{Video quality issue distribution in over 1M annotated clips used for training the quality model.}
        \label{fig:video_quality}
    \end{subfigure}
    \hfill
    \begin{subfigure}{0.48\textwidth}
        \centering
        \includegraphics[width=\linewidth]{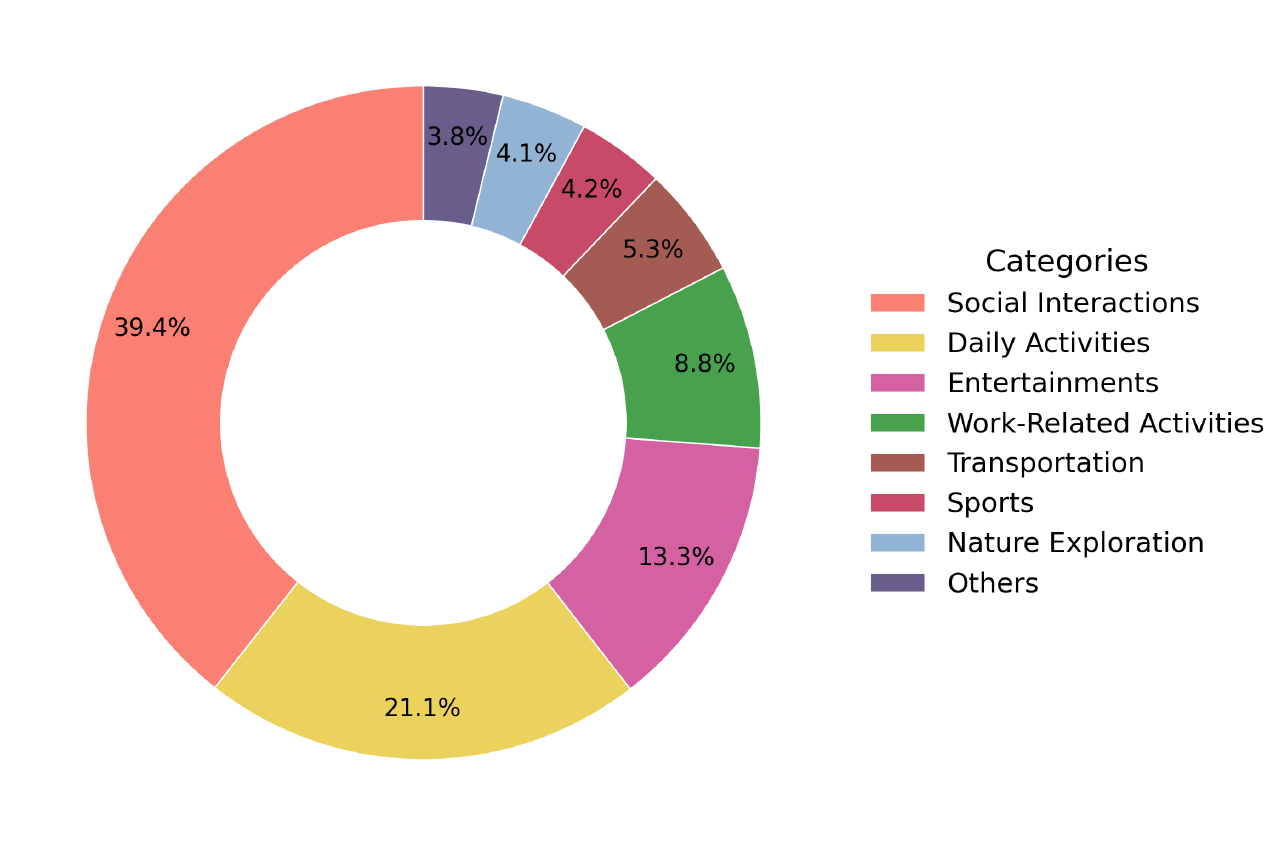}
        \caption{Action label distribution in the pre-training data, dominated by Social Interactions and Daily Activities.}
        \label{fig:action-type}
    \end{subfigure}

    \caption{Overview of the data filtering process and dataset characteristics. }
    \label{fig:data-overview}
\end{figure}

Quality scoring has certain limitations. For example, it is difficult for the motion score to accurately evaluate slow-motion videos; the aesthetics score may not be reliable within certain score ranges; and OCR filtering can miss some watermarks or small text. To obtain higher-quality training data after quality scoring, we train a quality model based on MLLM to further filter the data. We select over 1 million video clips with relatively high quality after quality scoring for manual annotation, labeling high-quality samples, and 13 different dimensions of low quality (see Fig.~\ref{fig:video_quality} for specific dimensions). We use this annotated data to perform SFT on VideoLLaMA3~\citep{zhang2025videollama}, supporting both landscape and portrait videos. During 480p training, we used this model to filter the training data. We also conduct a manual evaluation of the quality model’s predictions on the validation set, finding that the accuracy for samples predicted as high quality was 78\%.

\subsection{Caption Model}
\label{sec:caption}
The caption comprehensively describes various elements, including actions, subjects, backgrounds, camera motions, style, spatial relationships, etc. Notably, the description of actions is articulated with exceptional detail. Precise and fine-grained action descriptions play a vital role in improving both motion fidelity and the instruction-following capabilities of video generation models with respect to detailed actions.

\textbf{Enhancing action temporal understanding.}
For each complete sub-action described in the annotated captions, we further annotate its start and end timestamps. We prompt the model with questions regarding the specific actions occurring within a given start and end time, and ask the model to respond by describing the sequence of sub-actions that take place during this interval. Through this data construction approach, the model is able to better align fine-grained action descriptions with corresponding video segments during training, thereby enhancing its capability to understand the temporal order of actions.

\textbf{Model Training.}
Based on the Qwen2.5-VL~\citep{bai2025qwen2}, we conduct joint training with both caption and sub-action description tasks. For the caption task, the model takes the Tariser2~\citep{yuan2025tarsier2} caption as input and generates detailed descriptive caption as output. The visual encoder is kept unfrozen during this stage. The training conducted during the DPO stage is aimed at mitigating hallucination. On the DPO~\citep{rafailov2024dpo} side, DPOP~\citep{pal2024dpop} loss is employed to maintain the stability of long-form outputs. In this stage, the visual encoder is frozen. The caption model is trained on English-language data.

\subsection{Semantic Balancing}
\label{sec:caption}

A balanced training dataset is crucial for model performance.
However, in practice, certain categories are severely underrepresented in our training data; for example, as shown in Figure \ref{fig:action-type}, data for the sports category is particularly scarce.
To address this, we constructed an action labeling system, annotated the training data according to this taxonomy, and balanced the data based on the distribution of these action labels.
Specifically, we use Qwen2.5-32B~\citep{qwen2.5} to perform multi-level action label classification on the caption of the training videos, which includes 12 first-level action labels, 100 second-level action labels, and over 6,000 third-level action labels.
We then analyzed the distribution of each level of action labels in the data.
If an action label's proportion was too low, we addressed this imbalance by oversampling existing data and generating additional synthetic samples.

\subsection{Hierarchical Data Filtering}
\label{sec:data_filtering}
Our data filtering strategy consists of several hierarchical stages, each applying stricter and more comprehensive quality criteria. Early stages focus on keeping diverse video content and basic semantic consistency. In later stages, we add more advanced checks for visual quality, composition, and data balance. By the final stage, only the highest-quality and high-resolution samples are kept. This process ensures the dataset is gradually refined to meet the needs of each training phase and supports effective model development. 

\begin{figure}[htbp]
    \centering
    \includegraphics[width=0.9\textwidth]{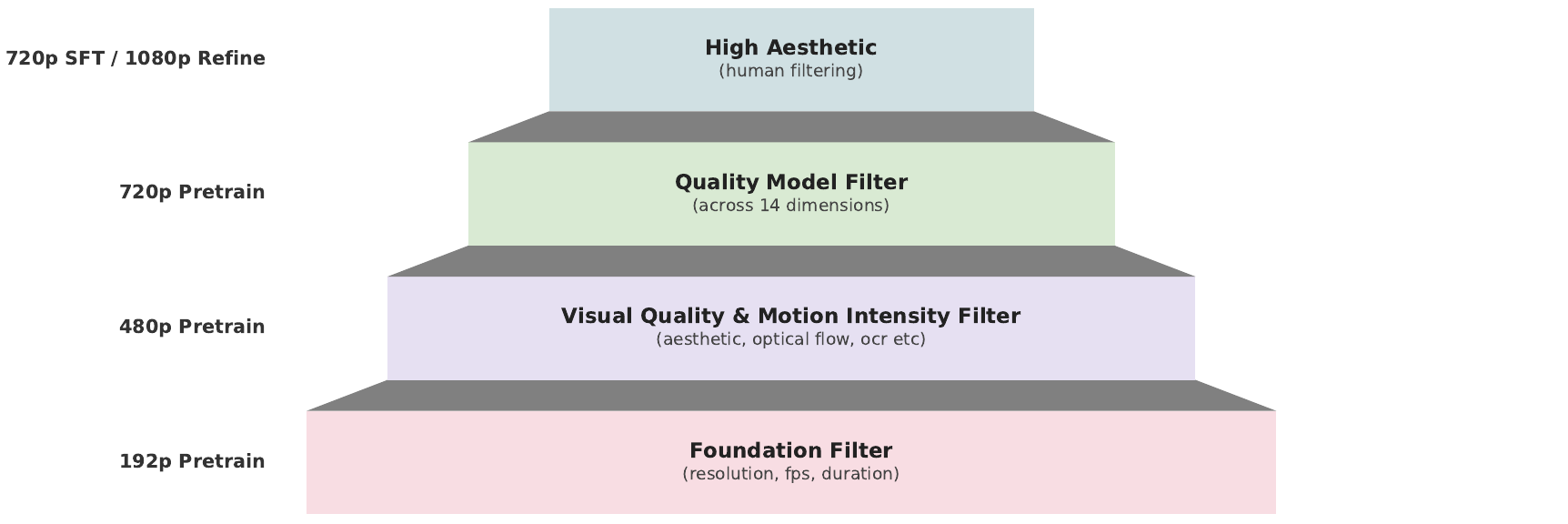}
    \caption{
        The hierarchical data filtering funnel, which progressively refines the dataset through increasingly strict quality criteria.
    }
    \label{fig:data-filter}
\end{figure}

\textbf{192p Pre-training.}
At this initial stage, we retain as much video data as possible to expose the model to diverse semantics and video types at low resolution. Filtering mainly removes clips with excessive overlaid text to avoid meaningless text displays, as well as static videos with minimal motion. Here, we focus on maintaining motion consistency, semantic alignment between text and video, and temporal coherence, without strict requirements on high-level visual quality.

\textbf{480p Pre-training.}
Building on the previous stage, this phase introduces aesthetic scoring, stricter thresholds for overlaid text and motion, and an 8-dimensional quality model. Clips are now filtered for overall visual quality, artifacts such as black borders, texts or watermarks, playback speed anomalies, perspective changes, and motion stability, enabling more effective removal of low-quality or problematic videos.

\textbf{720p Pre-training.}
While basic scoring and filtering remain similar to the previous stage, this stage expands the quality model to 12 dimensions by adding four new dimensions: subject cutoff, overexposed, dart or blurry, and image animation to better capture composition and color issues. This enables a more refined selection of high-quality clips.

\textbf{720p Continue Training (CT).}
At this stage, we further enhance video quality by adding two new criteria: video definition and color balance. Additionally, we enforce balance across different data sources and semantic categories to ensure diversity and representativeness, even as the dataset size is further reduced through stricter filtering.

\textbf{720p Supervised Fine-Tuning (SFT).}
At this stage, we introduce synthetic data, including both highly aesthetic and surreal video samples. Although the training target remains 720p, only high-resolution (2k/4k) real video clips are retained to ensure maximum clarity. For complex motion videos, manual selection and detailed caption annotation are performed to further improve data quality.

\textbf{1080p Refinement.}
At this final stage, we focus on maximizing video clarity. Only the 720p CT data and the highest-quality 2k/4k high-resolution data are retained, providing the best possible samples for advanced model training.


%% file: content/4_recipe.tex
\section{Training / Inference Recipe}

In this section, we provide a detailed introduction to the entire training and inference process of the model, as well as how to improve the generated video's motion, aesthetics, realism, and color from the perspectives of data, training, noise scheduling, hyperparameter tuning, and inference.

\subsection{Multi-stage Training}
Our entire multi-stage training schedule is shown in Tab.~\ref{tab:training_stages}, which contains the tasks of T2I, T2V, I2V, and Refiner. We also present the data volume of each stage and the corresponding training hyperparameters.

\textbf{T2I. } 
We first train the T2I task, starting with 256p images and progressively increasing the resolution up to 1024p, covering multiple aspect ratios such as 1:1, 16:9, 9:16, 4:3, and 3:4. The T2I task enables the model to learn the correspondence between text and images and enhance the model's instruction following capability.

\textbf{T2V \& I2V. }
After T2I, we begin training the T2V task, where the video aspect ratio is consistently maintained at 16:9 and 9:16. We also include T2I in joint training, starting from 192p, with the video resolution set to 192×352, and both images and videos kept at the same resolution. We first train at 192p at a frame rate of 12fps, followed by 192p 16fps. We find that using a lower fps helps the model better learn motion in videos. Starting from 480p, we jointly train the T2V and I2V tasks by controlling whether to include the first-frame condition, with a 20\% probability of training I2V. The video resolution is 480×864 at a frame rate of 16fps. Training at 720p is divided into three stages: Pretraining, Continue Training (CT), and Supervised Finetuning (SFT), with gradually improving data quality. For T2I, in addition to 720×1280, we also add 1080×1920 image data for training. Our experiments show that including 1080p images in training can improve visual quality. 

\textbf{Refiner. }
The 1080p training for the Refiner is based on the checkpoint after 720p CT, using both 1080×1920 video and image data, and is also divided into Refine and SFT stages. During the SFT stage, we use only images and videos with resolutions above 2K as training data to improve the overall visual quality.

\textbf{Training hyperparameters. }
Throughout the entire training process, we ensure that the data from each stage is seen for at least one epoch, and the overall learning rate shows a decreasing trend. Following SD3~\citep{esser2024sd3}, the sigma shift in flow matching increases as the resolution increases. 

\begin{table}[!t]
\centering
\renewcommand{\arraystretch}{1.3} 
\small
\resizebox{\textwidth}{!}{
\begin{tabular}{l|ccccccc}
\textbf{Task} & \textbf{Stage} & \textbf{Resolution} & \textbf{Data volume} & \textbf{Epochs} & \textbf{Learning rate} & \textbf{Sigma shift} \\
\shline
T2I & Pretrain & 256p  & 170M  & 24 & 1e-4 & 1.0 \\
T2I & Pretrain & 512p  & 100M  & 3 & 6e-5 & 2.0 \\
T2I & Pretrain & 1024p  & 60M  & 1 & 3e-5 & 3.0 \\
\shline
T2V / T2I & Pretrain & 192p 12fps  & 185M / 40M & 2 & 1e-4 & 1.5 \\
T2V / T2I & Pretrain & 192p 16fps  & 185M / 40M & 1.5 & 1e-4 & 1.5 \\
T2V+I2V / T2I & Pretrain & 480p 16fps  & 41M / 40M & 3 & 5e-5 & 2.5 \\
T2V+I2V / T2I & Pretrain & 720p 24fps  & 22M / 15M & 1 & 3e-5 & 3.0 \\
T2V+I2V / T2I & CT & 720p 24fps  & 4M / 15M & 2 & 1e-5 & 3.0 \\
T2V+I2V / T2I & SFT & 720p 24fps  & 0.1M / 15M & 2 & 5e-6 & 3.0 \\
\shline
(T+V)2V / T2I & Refine & 1080p 24fps  & 3M / 3M & 1 & 1e-5 & 4.5 \\
(T+V)2V / T2I & SFT & 1080p 24fps  & 0.2M / 3M & 1 & 5e-6 & 4.5 \\
\end{tabular}
}
\caption{The entire training process for the four tasks: T2I, T2V, I2V, and Refiner. In the ``Data volume'' column, the value before the ``/'' refers to the amount of video data, while the value after the ``/'' refers to the amount of image data. ``M'' refers to million. }
\label{tab:training_stages}
\end{table}

\subsection{Representation Alignment}
Motivated by recent advances in image generation, where studies like REPA~\citep{yu2024representation} and VA-VAE~\citep{yao2025reconstruction} have established that leveraging semantic information significantly improves model convergence, we investigate a parallel approach for video generation.
We hypothesize that integrating video-level semantic information can similarly accelerate the training process of video generation models.

To this end, we employ Qwen2.5-VL~\citep{bai2025qwen2} to extract high-level semantic features from videos, a model chosen for its capability to process inputs of arbitrary length. 
These semantic features are then aligned with the intermediate features of the DiT model by applying a constraint based on cosine similarity.
For computational efficiency, this constraint is selectively applied only during the 480p training stage for both video and image data. 
This avoids impeding the fast-iterating 192p stage and circumvents the prohibitive storage overhead required for 720p features.

\begin{figure}[!t]
    \centering
    \includegraphics[width=1.0\textwidth]{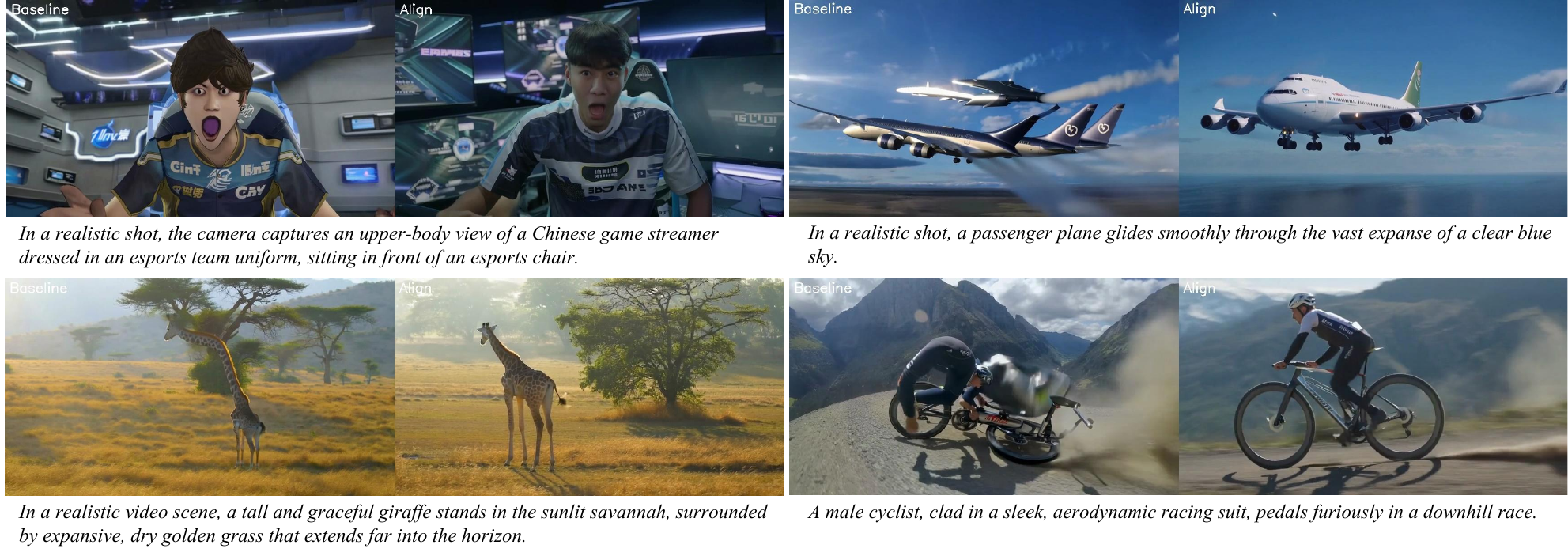}
    \caption{
        Qualitative comparison of generated videos with and without our representation alignment method. For each of the four cases shown, the frame on the left is from the baseline model (no alignment), while the frame on the right is from our model with the alignment constraint.
    }
    \label{fig:repa}
\end{figure}

Concretely, the alignment operates on two feature sets.
The first, denoted as $\textbf{h} \in \mathbb{R}^{H_h \times W_h \times T_h \times C_h}$, are latent features extracted from the 16th layer of our Task-Unified DiT model, where $H, W, T$ and $C$ represent the height, width, time, and channel dimensions, respectively.
The second, $\textbf{f} \in \mathbb{R}^{H_f \times W_f \times T_f \times C_f}$, are the corresponding high-level semantic features from Qwen2.5-VL.
To optimize training throughput, $\textbf{f}$ are pre-computed and stored offline.
To harmonize these features for alignment, we perform two preprocessing steps on the DiT features $\textbf{h}$.
First, to mitigate storage pressure and match spatial-temporal dimensions, we downsample $\textbf{h}$ to produce $\textbf{h}_d = d(\textbf{h}) \in \mathbb{R}^{H_f \times W_f \times T_f \times C_h}$.
This involves a downsampling factor of 2 in spatial dimensions and 4 in the temporal dimension (i.e., $H_h = 2H_f, W_h = 2W_f, T_h=4T_f$).
Second, we employ a lightweight MLP, $g_{\phi}$, to project the channel dimension of the downsampled features, yielding $\textbf{h}_g = g_{\phi}(\textbf{h}_d)$, which now shares the same channel dimension as $\textbf{f}$.
The alignment is then achieved by maximizing the cosine similarity between the processed DiT features $\textbf{h}_g$ and the semantic features $\textbf{f}$. This is formulated as the following minimization objective:
$$
    \mathcal{L}_{\text{align}} = -\mathbb{E}_{\mathbf{x}}\left[\frac{1}{N}\sum_{i=1}^{N}\frac{\textbf{h}_{g,i} \cdot \textbf{f}_i}{\lVert\textbf{h}_{g,i}\rVert \lVert\textbf{f}_i\rVert}\right]
$$
where $\mathbf{x}$ represents the training data, and the sum is over all $N=H_fW_fT_f$ feature vectors in the tensor. 
Finally, this alignment loss is incorporated into our final training objective:
$$
\mathcal{L}_{\text{train}} = \mathcal{L}_{\text{fm}} + \lambda \mathcal{L}_{\text{align}},
$$
where $\mathcal{L}_{\text{fm}}$ is the primary flow matching objective and $\lambda$ is a hyperparameter, set to 0.5 in our experiments.
The inclusion of representation alignment constraint yields a significant enhancement in the semantic quality of the generated videos.
We demonstrate this through an ablation study comparing our method against a baseline trained without the constraint. 
Both models were initialized from an identical 192p checkpoint and trained for 12,000 steps at 480p. 
Fig.\ref{fig:repa} provides a side-by-side comparison, where the outputs from our method clearly display more organized and meaningful semantic structures.

\subsection{Motion Optimization}
We optimized motion for the T2V and I2V tasks from three aspects: low-resolution video training, noise scheduling, mixed training of T2V \& I2V, and poor motion data filtering. These improvements enable \method to generate videos with large-scale, smooth motion.

\begin{figure}[!t]
    \centering
    \includegraphics[width=1.0\textwidth]{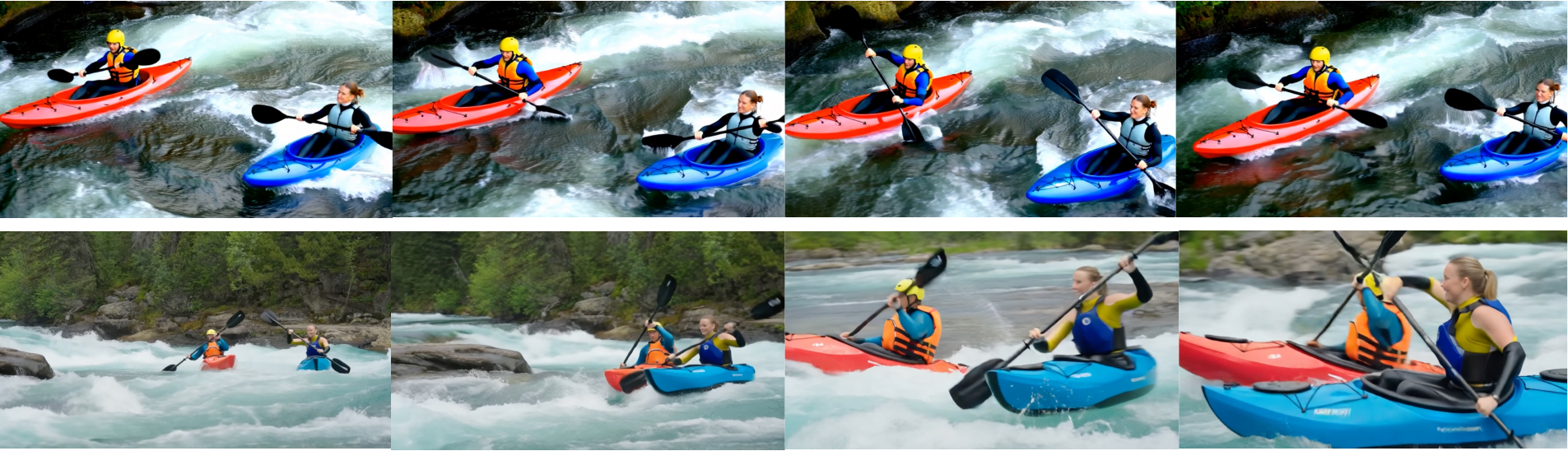}
    \caption{
        \textbf{Top:} 480p T2V results without 192p pretrain. \textbf{Bottom:} 480p T2V results with 192p pretrain. Results for prompt ``Two people are paddling two kayaks vigorously in the river.''
    }
    \label{fig:low_res_motion}
\end{figure}

\textbf{Low resolution video training. }
We find that training with low-resolution videos (such as 192p) is crucial for the model to learn motion. \method learns motion at 192p, dividing the process into two stages: 12fps and 16fps. This approach helps to partially decouple motion learning from visual quality and semantics. Compared to directly switching from the T2I training to 480p T2V training, we find that introducing 192p T2V training as an intermediate step enables faster convergence during 480p training and results in generated videos with larger motion, as is shown in Fig.~\ref{fig:low_res_motion}. Therefore, we invest considerable data and computational resources in 192p T2V training.

\textbf{Noise scheduling. }
We follow the flow matching schedule proposed in SD3~\citep{esser2024sd3}. It highlights that timestep sampling is crucial for flow matching training. Specifically, timesteps near the middle of the [0, 1] interval are more challenging for the model to learn. Therefore, probability density functions that assign higher probability to the middle and lower probability to the ends, such as the logit-normal and mode functions, are used for timestep sampling, as is shown in Eq.~\ref{eq:logitnormal} and Eq.~\ref{eq:modesampler}, respectively. 

\begin{equation}
\pi_{\text{ln}}(t; m, s) = \frac{1}{s\sqrt{2\pi}} \frac{1}{t(1-t)}\exp\Bigl(-\frac{(\text{logit}(t)-m)^2}{2s^2}\Bigr),
\label{eq:logitnormal}
\end{equation}

\begin{equation}
f_{\text{mode}}(u; s) = 1 - u - s\cdot\Bigl(\cos^2\bigl(\frac{\pi}{2} u\bigr) - 1 + u\Bigr).
\label{eq:modesampler}
\end{equation}

\begin{figure}[!t]
    \centering
    \includegraphics[width=1.0\textwidth]{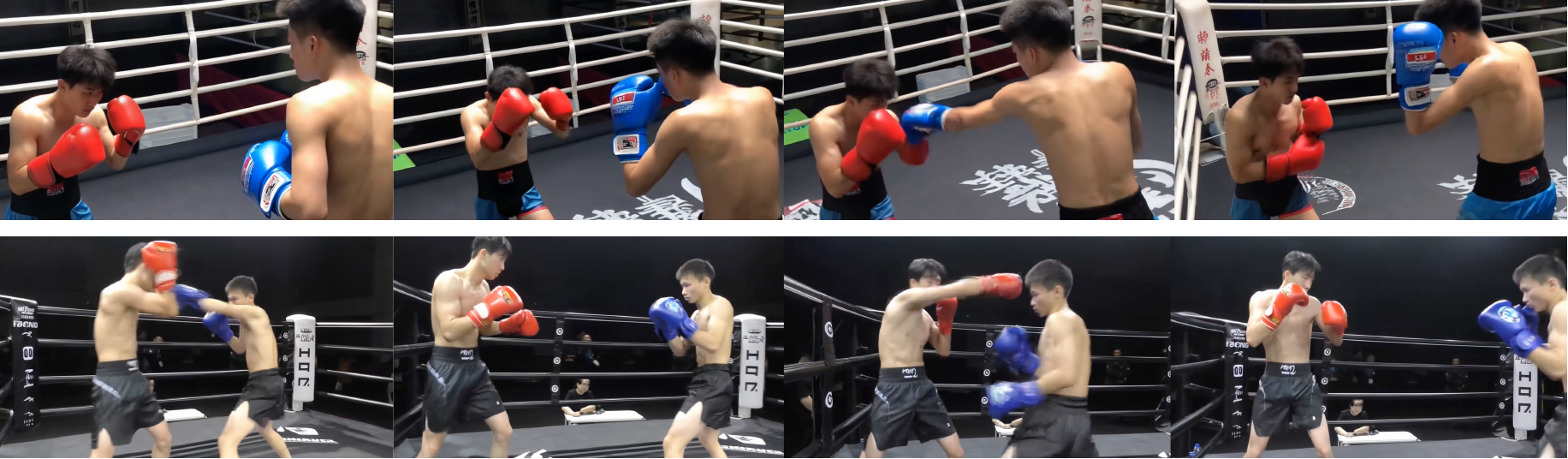}
    \caption{
        \textbf{Top:} 720p T2V results with logit-normal sampling. \textbf{Bottom:} 720p T2V results with mode sampling. Results for prompt ``In the center of the boxing ring, two male boxers are fighting each other.''
    }
    \label{fig:noise_case}
\end{figure}

\begin{figure}[!t]
    \centering
    \includegraphics[width=1.0\textwidth]{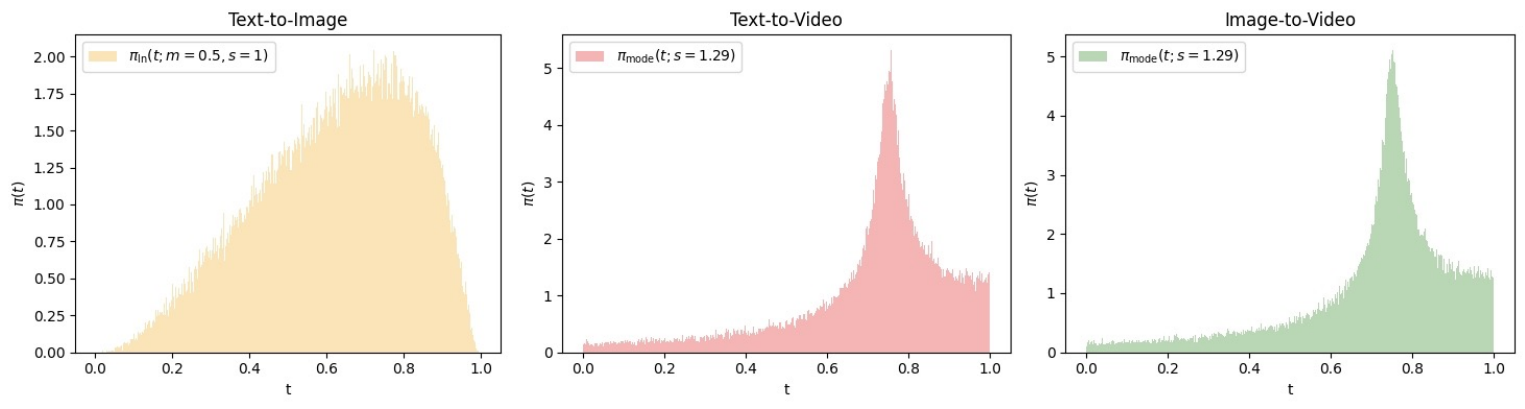}
    \caption{
        \textbf{Probability density functions for timestep sampling.}
        While Text-to-Image uses a broad logit-normal distribution (Eq.~\ref{eq:logitnormal}), Text-to-Video and Image-to-Video tasks employ a sharply peaked mode distribution (Eq.~\ref{eq:logitnormal}) to generate larger motion amplitudes.
    }
    \label{fig:noise}
\end{figure}

We conduct ablation studies on different timestep sampling functions on 192p, 480p, and 720p T2V tasks, and the results are consistent across settings. We find that mode sampling enables the model to generate videos with larger motion amplitudes, as is shown in Fig.~\ref{fig:noise_case}. Both the top and bottom models were trained for 10,000 steps under the 720p setting. The top model uses logit-normal sampling, while the bottom model uses mode sampling. We observe that the model with mode sampling generates videos in which the two boxers are engaged in a more intense fight, with more frequent and faster actions. So we adopt Lognorm(0.5, 1) for the T2I task, Mode(1.29) for the T2V and I2V tasks. The distribution of timestep sampling is shown in Fig.~\ref{fig:noise} with sigma shift = 3.

\textbf{T2V \& I2V joint training. }
Motion is a crucial optimization objective for the I2V task, as all generated sequences share the same initial frame, and the quality of the subsequent motion largely determines the overall performance of the model. The I2V task often suffers from the issue of limited motion, as the strong conditioning on the initial frame leads the model to generate subsequent frames that are overly similar to the first frame. Some studies~\citep{liu2025dynamic,tian2025extrapolating,zhang2025motionpro} address this problem by employing sophisticated designs, such as incorporating external supervision, representation, or conditioning, to enhance the motion generated by the model. 

\begin{figure}[!t]
    \centering
    \includegraphics[width=1.0\textwidth]{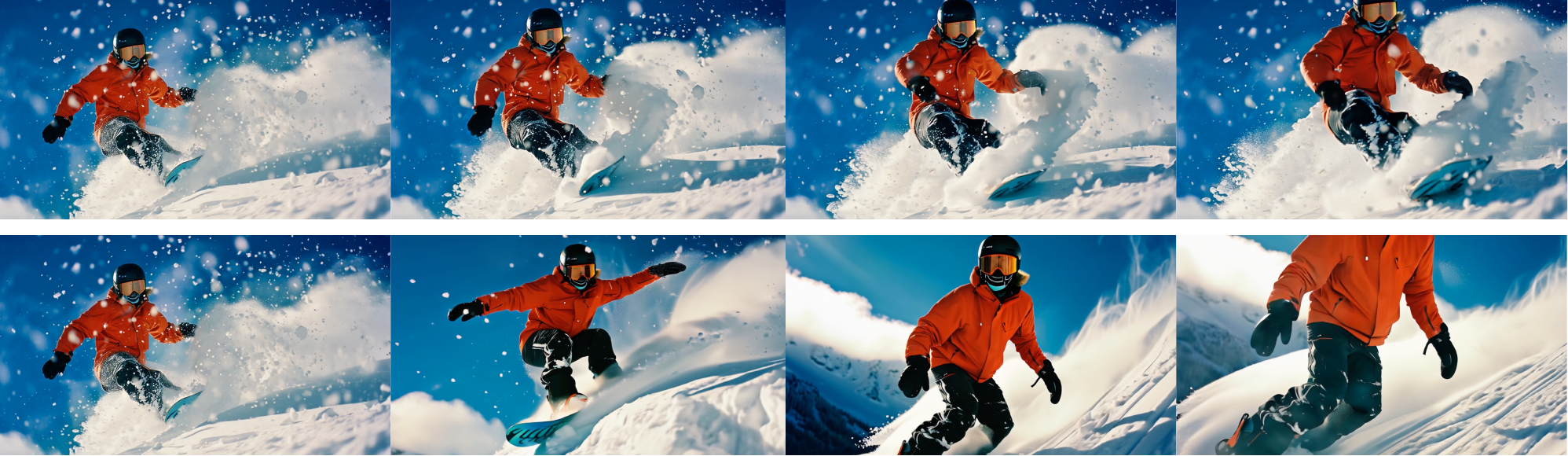}
    \caption{
        \textbf{Top:} 720p I2V results without T2V joint training. \textbf{Bottom:} 720p I2V results with T2V joint training. Results for prompt ``A man is skiing.''
    }
    \label{fig:i2v_show}
\end{figure}

We adopt a simpler approach to enhance motion in the I2V task by jointly training I2V and T2V. Specifically, we introduce the initial frame as a condition for the I2V task with a probability of 20\%. This strategy prevents the initial frame conditioning from becoming overly dominant and helps maintain comparable motion magnitudes between I2V and T2V. In the early stages of training, we observe that the initial frame consistency in I2V is relatively poor; however, as training converges, both the initial frame consistency and the magnitude of motion in I2V improve significantly. We compare the motion generated by models jointly trained on T2V and I2V tasks with that trained solely on I2V. As shown in Fig.\ref{fig:i2v_show}, the results clearly demonstrate that joint training leads to significantly larger motion amplitudes in the I2V task.

\begin{figure}[!t]
    \centering
    \includegraphics[width=1.0\textwidth]{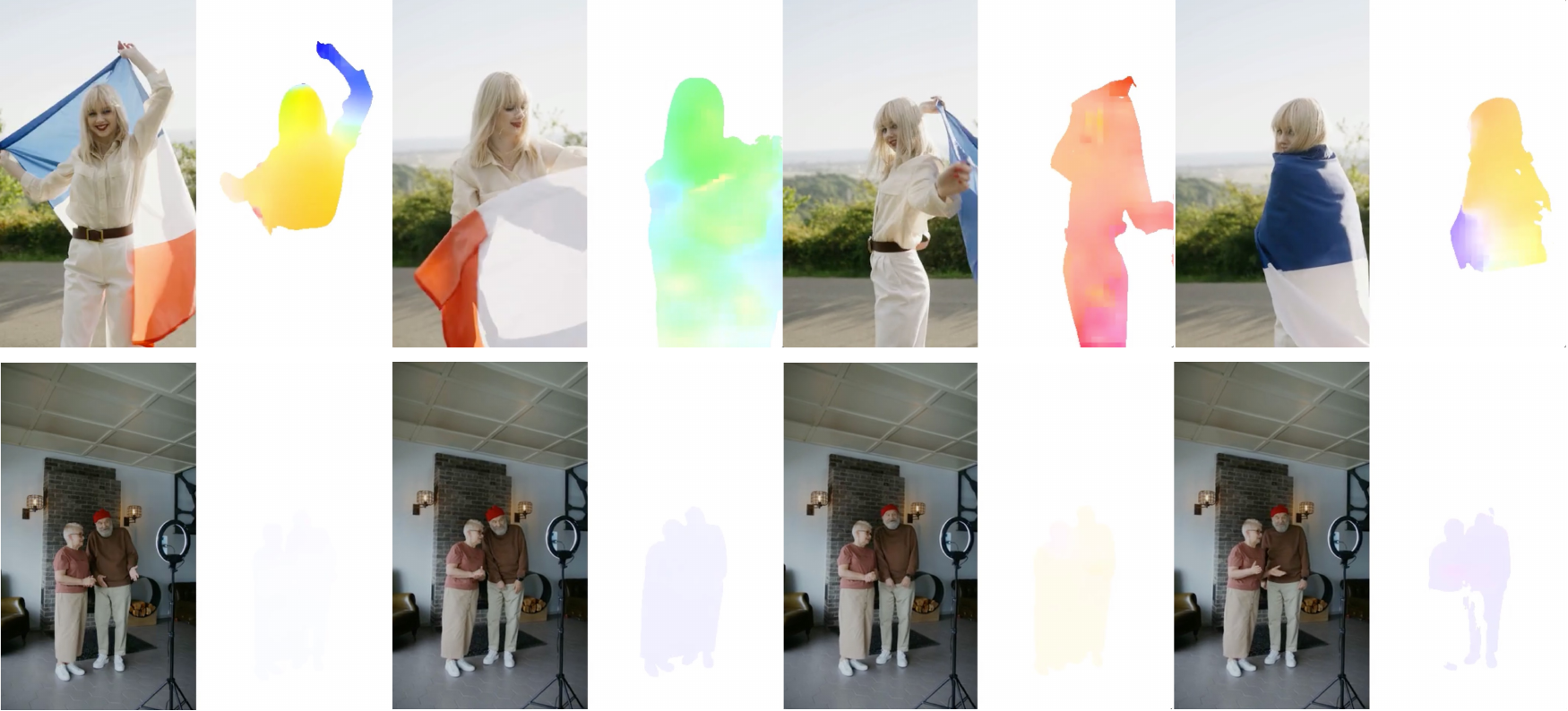}
    \caption{
        Raw training video clips and the foreground optical flow. \textbf{Top:} Foreground motion score is 11.0. \textbf{Bottom:} Foreground motion score is 0.75.
    }
    \label{fig:motion_score}
\end{figure}

\textbf{Poor motion data filtering. }
The motion amplitude in the training data has a critical impact on the motion quality of generated videos. We primarily filter out static videos and those with excessive motion using a foreground motion score mentioned in Sec.~\ref{sec:data_filtering}. Fig.~\ref{fig:motion_score} presents examples of videos with different foreground motion scores. However, we found that some slow-motion and image-effect videos cannot be effectively filtered by the motion score alone. These two types of videos can also significantly degrade the model's motion quality. Therefore, we further filter out slow-motion and image-animation videos using the video quality model described in Sec.\ref{sec:quality_model}. After these two filtering steps, the remaining videos generally exhibit motion within a normal range and are suitable for training models to achieve better motion performance.

\subsection{Aesthetics Optimization}
Obtaining high-aesthetic video data is significantly more challenging than collecting high-quality images. This gap is particularly pronounced for dynamic, high-motion scenes, as real-world footage is often sourced from documentary-style recordings which prioritize capturing an event over artistic composition and are susceptible to technical flaws like motion blur. The challenge of generating coherent, high-fidelity motion is a primary focus of state-of-the-art video generation research. To address this critical data gap, we introduce a dedicated synthetic data enhancement stage~\cite{girdhar2023emu}, aimed at further boosting both the visual quality and creative capacity of our model by leveraging high-quality image assets. 

\begin{figure}[ht]
    \centering
    \includegraphics[width=1.0\textwidth]{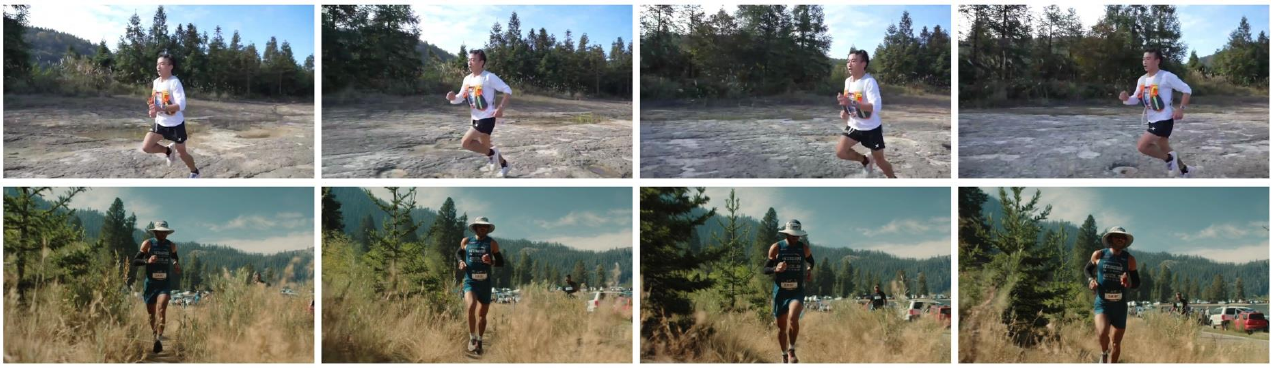}
    \caption{
        Aesthetic comparison for a running scene. \textbf{Top}: Typical real-world video. \textbf{Bottom}: Our synthetic data shows superior visual quality, including better lighting and composition.
    }
    \label{fig:data_comparison}
\end{figure}

\textbf{Step 1: Synthetic Data Generation.}
Leveraging our unified multi-task model, we utilize high-quality image datasets and synthesized surreal images, along with our I2V capability, to construct a diverse collection of synthetic video samples. These videos are meticulously generated to emphasize aesthetic value, featuring visually striking scenes, advanced composition, and sophisticated use of color and lighting. This strategy also enhances the model’s ability to follow creative instructions and handle surreal concepts. As illustrated in Figure~\ref{fig:data_comparison}, our synthetic data exhibits a significant improvement in aesthetic quality over typical real-world footage.

\begin{figure}[t]
    \centering 

    \begin{subfigure}[b]{0.48\textwidth}
        \centering
        \includegraphics[width=\textwidth]{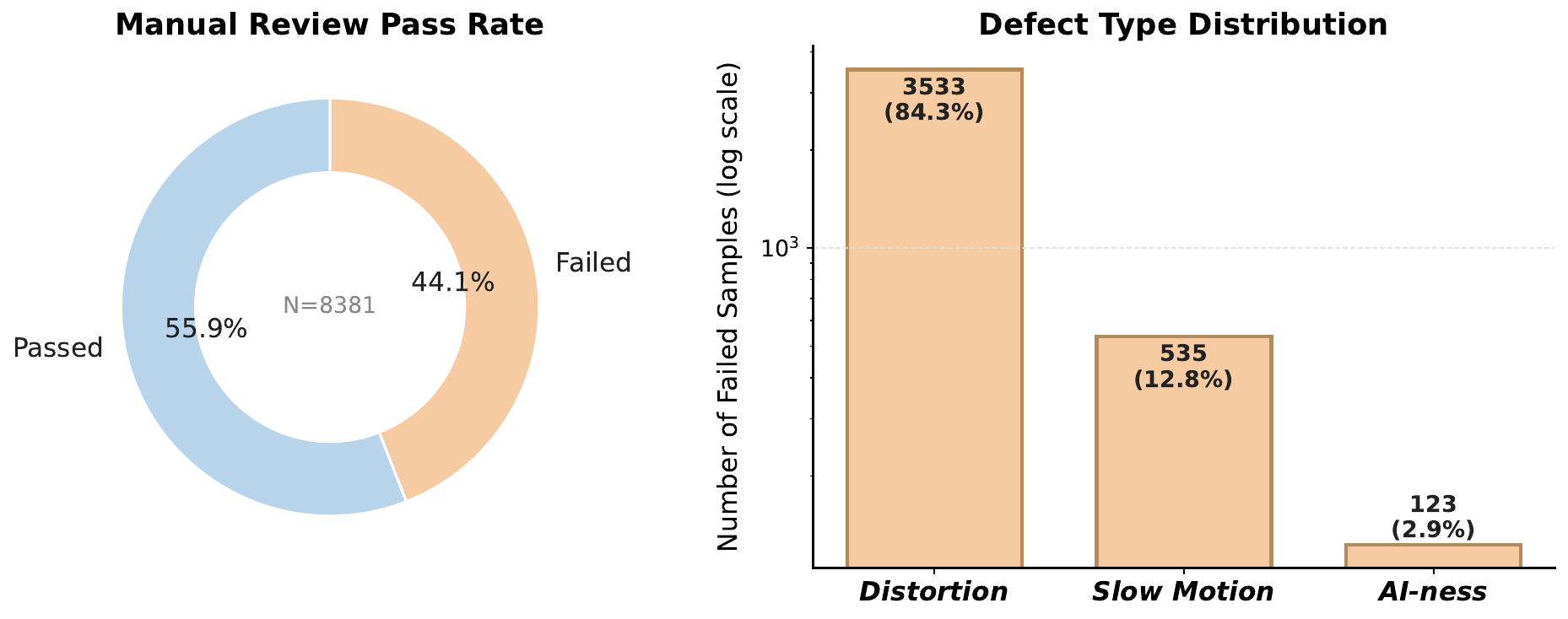}
        \caption{
            Manual review statistics for synthetic video samples (N=8381), where distortion is the primary failure reason
        }
        \label{fig:review_stats} 
    \end{subfigure}
    \hfill 
    \begin{subfigure}[b]{0.48\textwidth}
        \centering
        \includegraphics[width=\textwidth]{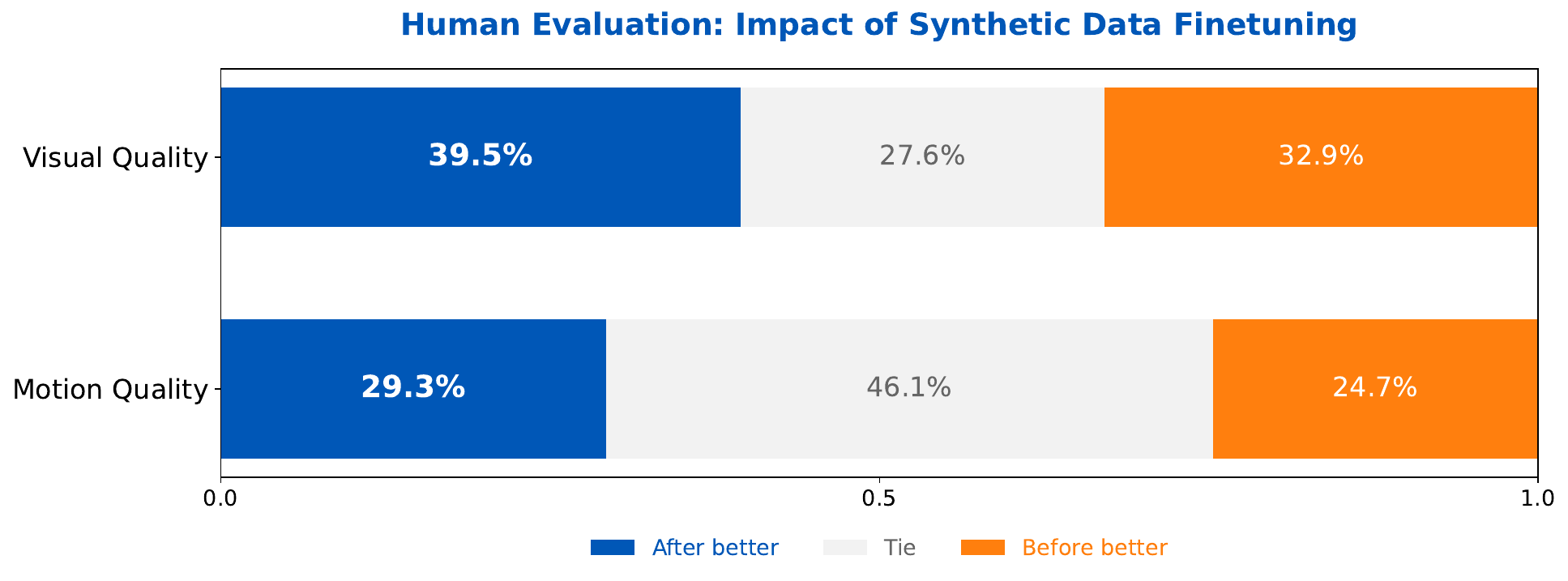}
        \caption{
            Human evaluation (SBS) results before and after high-aesthetic finetuning, which shows a ~7\% improvement in visual quality after finetuning on the curated synthetic data, with motion quality preserved.
        }
        \label{fig:sbs_finetune} 
    \end{subfigure}

    \caption{
        Statistics and impact of our synthetic data enhancement stage. 
    }
    \label{fig:synthetic_data_evaluation} 
\end{figure}

\textbf{Step 2: Rigorous Quality Control.}
To ensure the highest quality, all synthetic videos undergo a rigorous manual review. Samples exhibiting distortions, excessive "AI-ness" (such as loss of fine details or unnatural saturation), or insufficient motion are filtered out. Only samples that pass this review are included in the training set. Each video clip is independently reviewed by at least two annotators, and only those with unanimous approval are retained. The statistics of this review process, including the pass rate and a breakdown of failure reasons, are detailed in Figure~\ref{fig:review_stats}.

\textbf{Step 3: Balanced Data Diet.}
We carefully balance the proportion of synthetic and real video data to preserve realism and prevent overfitting to synthetic styles. Our experiments indicate that relying solely on synthetic data can lead to loss of visual detail and limited semantic alignment, particularly given the limited scale of manually reviewed synthetic samples.

\begin{figure}[!t]
    \centering
    \includegraphics[width=1.0\textwidth]{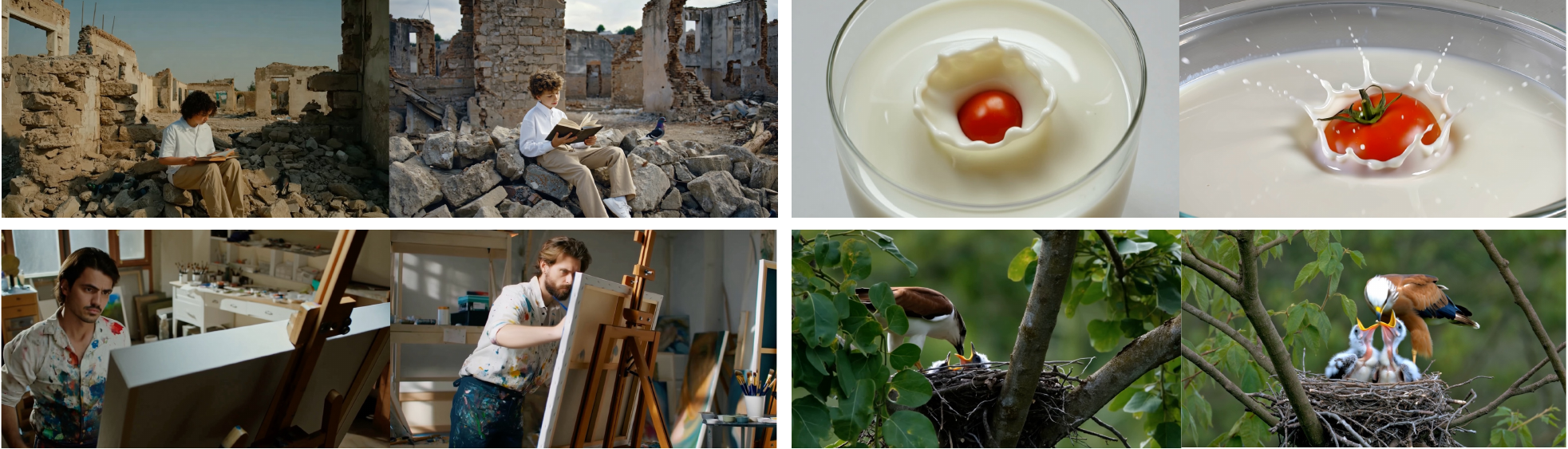}
    \caption{
        Comparison of four videos generated by models before and after high-aesthetic finetuning. For each case, the frame on the \textbf{left} is before finetuning, and that on the \textbf{right} is after finetuning. 
    }
    \label{fig:aes_show}
\end{figure}

\textbf{Finetuning Results.}
 We conduct high-aesthetic finetuning using this curated data. This aesthetics optimization strategy yields significant improvements in visual quality for the T2V task, without compromising temporal coherence. This is validated through both quantitative human evaluations and qualitative analysis. 
 \textbf{Quantitative improvements} are demonstrated in a side-by-side human evaluation on a 304-prompt general benchmark (Figure~\ref{fig:sbs_finetune}). The fine-tuned model achieved a \textbf{39.5\% win rate} in visual quality, substantially outperforming the base model's 32.9\%. Crucially, the evaluation also confirmed that motion quality remains comparable between the two models, indicating our approach successfully isolates and enhances aesthetic performance.
 \textbf{Qualitative enhancements} are clearly visible in generated samples (Figure~\ref{fig:aes_show}). After fine-tuning, videos exhibit noticeable improvements in color vibrancy, brightness, clarity, and overall chromatic harmony.

\subsection{Model Balancing}
The performance of video generation models is often characterized by various trade-offs. For example, from the perspective of model behavior, increased motion typically leads to greater blurriness, resulting in lower visual quality, while videos with high visual fidelity often exhibit limited motion. Moreover, larger motion is frequently accompanied by more artifacts, and there is also a trade-off between realism and the presence of artifacts. 

In terms of scene generalization, we observe that, especially in the early stages of training, our model often struggles to distinguish between cartoon and realistic styles, or it may perform well on realistic scenes but poorly on non-realistic ones, or vice versa—generating plausible non-realistic scenes that nevertheless exhibit an artificial "AI-like" appearance. These issues are particularly pronounced in the T2V task. Balancing different aspects of model performance remains a significant challenge. To address this, we employ strategies including prompt tagging, video APG, and model averaging.

\textbf{Prompt tagging. }
We employ a prompt tagging approach to distinguish between different types of training data. Our training dataset is sourced from a wide variety of origins, often exhibiting diverse styles and quality levels. We assign distinct tags to the training data based on both the data source and the quality model described in Sec.~\ref{sec:quality_model}.
In terms of style, we differentiate between categories such as anime (2D, 3D, etc.), game, CG, real, and synthetic data. During training, we prepend the caption with a prompt describing the video's style. Regarding quality, we distinguish between high-definition, slow-motion data and low-definition, fast-motion data, appending a prompt describing video quality to the end of the training caption. During inference, we incorporate prompts describing undesirable qualities, such as low definition or slow motion, into the negative prompt. For specific style requirements (e.g., anime style), we prepend the corresponding descriptive prompt to the overall prompt using prompt rewriting techniques.

\begin{figure}[ht]
    \centering
    \includegraphics[width=1.0\textwidth]{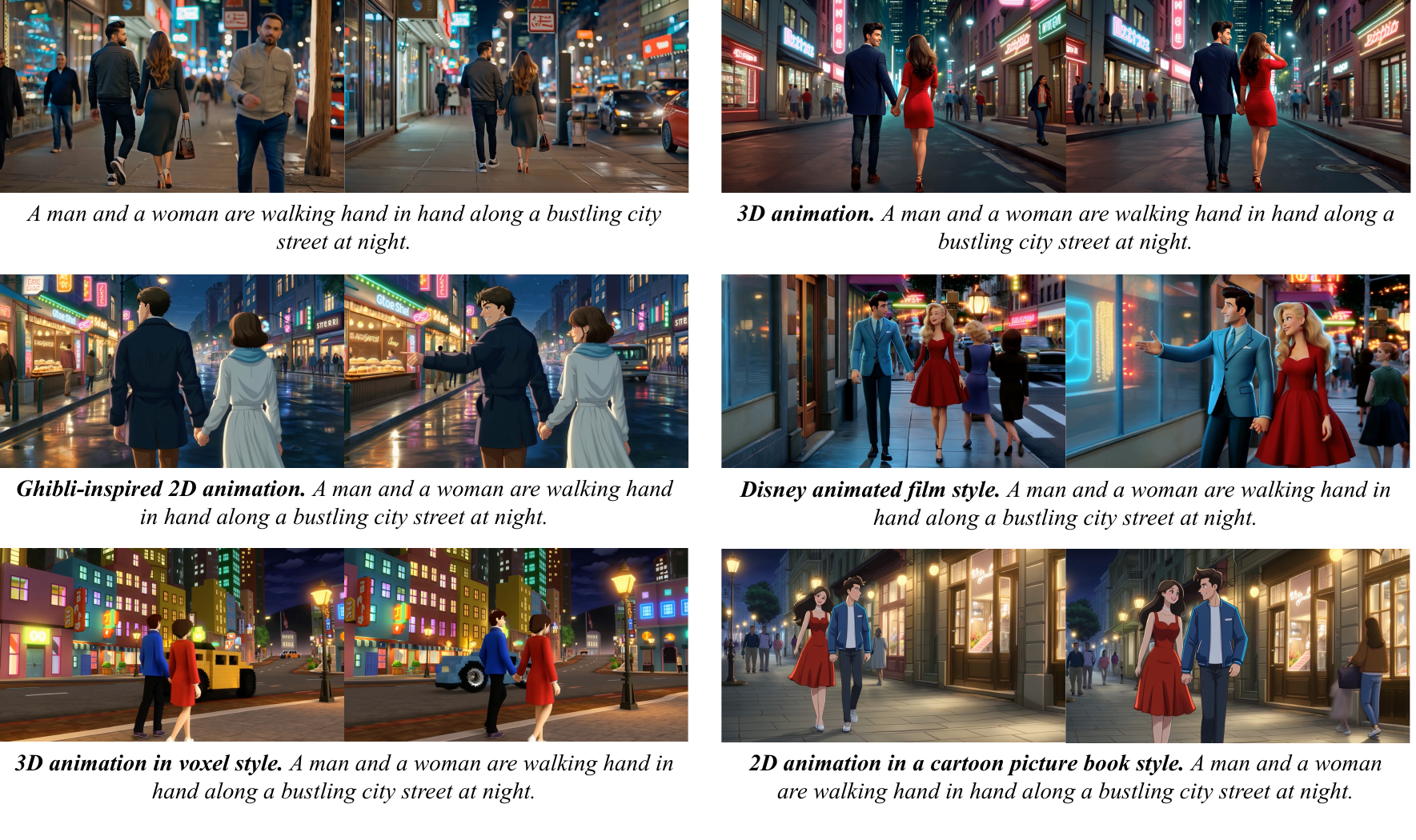}
    \caption{
        Six different styles of videos generated by \method T2V after prompt tagging. 
    }
    \label{fig:animation}
\end{figure}

\begin{figure}[htbp]
    \centering
    \includegraphics[width=1.0\textwidth]{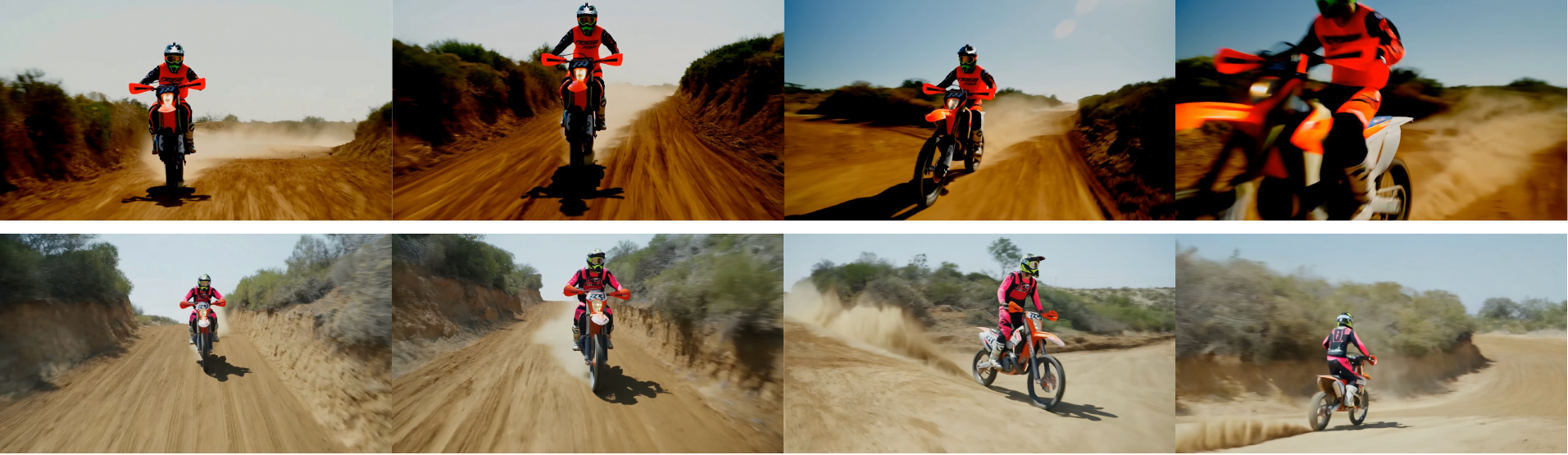}
    \caption{
        \textbf{Top:} 720p T2V results with CFG=5. \textbf{Bottom:} 720p T2V results with the optimal APG hyperparameters (CFG=8, norm threshold=27). Results for prompt ``An off-road motorcyclist speeds along a dusty motocross track, with the tires kicking up dirt and gravel.''
    }
    \label{fig:apg}
\end{figure}

\textbf{Video APG. }
We extend APG~\citep{sadat2024eliminating} to video generation to enhance realism and reduce artifacts. APG decomposes the update term in CFG into parallel and orthogonal components and down-weights the parallel component to achieve high-quality generations without oversaturation. We find that normalizing the latent from [C, H, W] dimension achieves fewer artifacts than from [C, T, H, W] dimension. For the hyperparameters, we find the normalization threshold 27 and guidance scale 8 achieve a good balance between realism and artifacts. Fig.~\ref{fig:apg} shows the comparison between CFG and video APG during inference. It is evident that video APG achieves greater realism without introducing additional artifacts.

\begin{figure}[ht]
    \centering
    \includegraphics[width=0.8\textwidth]{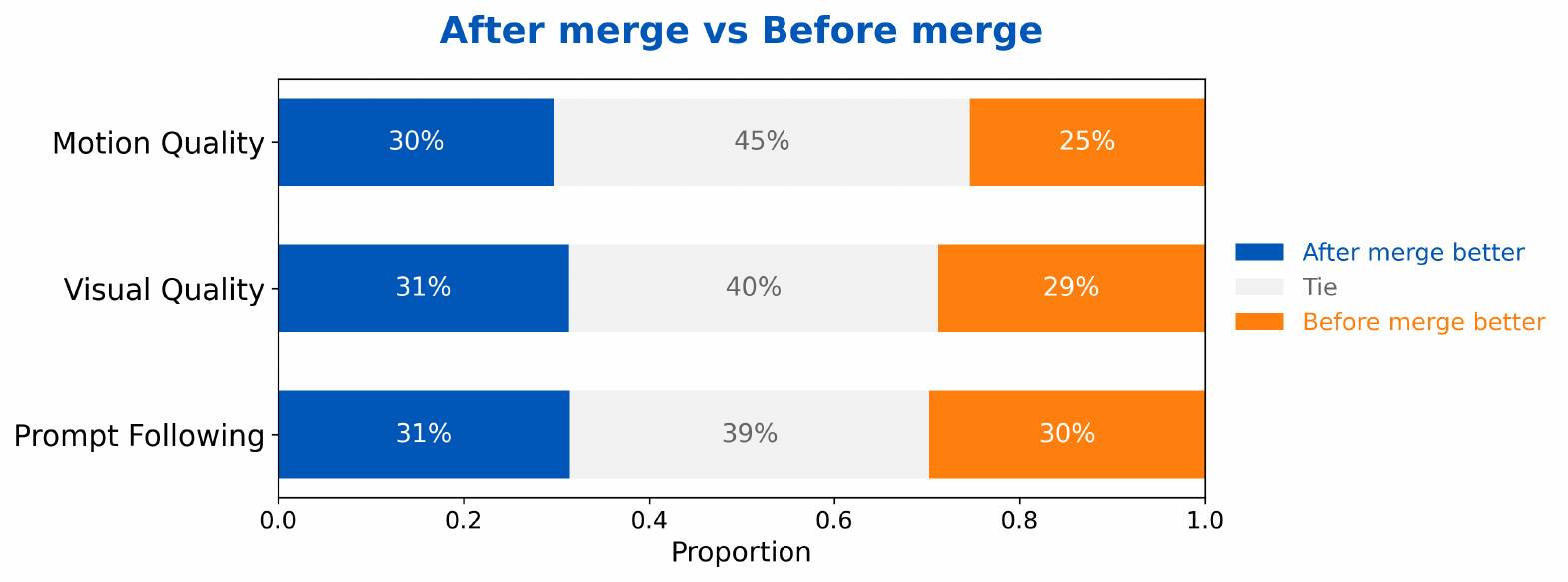}
    \caption{
        Human evaluation (SBS) results before and after model merging. Motion quality improves by 5\% after merging, while visual quality and prompt following improves by 2\% and 1\%.
    }
    \label{fig:sbs_merge}
\end{figure}

\textbf{Model averaging. }
We observe that models trained on different datasets, at various training stages, or with different timestep sampling probabilities exhibit distinct characteristics. For example, some models generate videos with excellent visual quality but slower motion, while others produce faster motion but with more artifacts. We employ a model averaging strategy~\citep{li2025model} to integrate these models with diverse attributes. Empirically, we find that merging 7 to 10 models yields favorable results. Comprehensive human evaluations presented in Fig.~\ref{fig:sbs_merge} indicate that the averaged model outperforms the individual model in visual quality, motion quality, and instruction following.

\subsection{Prompt Rewriting}
The purpose of prompt rewriting is to align diverse user inputs as closely as possible with the captions used during model training, thereby achieving results that are more consistent with the model’s training performance. We do not train a dedicated prompt rewriting model; instead, we directly utilize existing LLMs such as GPT-4.1. 

\begin{figure}[htbp]
    \centering
    \includegraphics[width=1.0\textwidth]{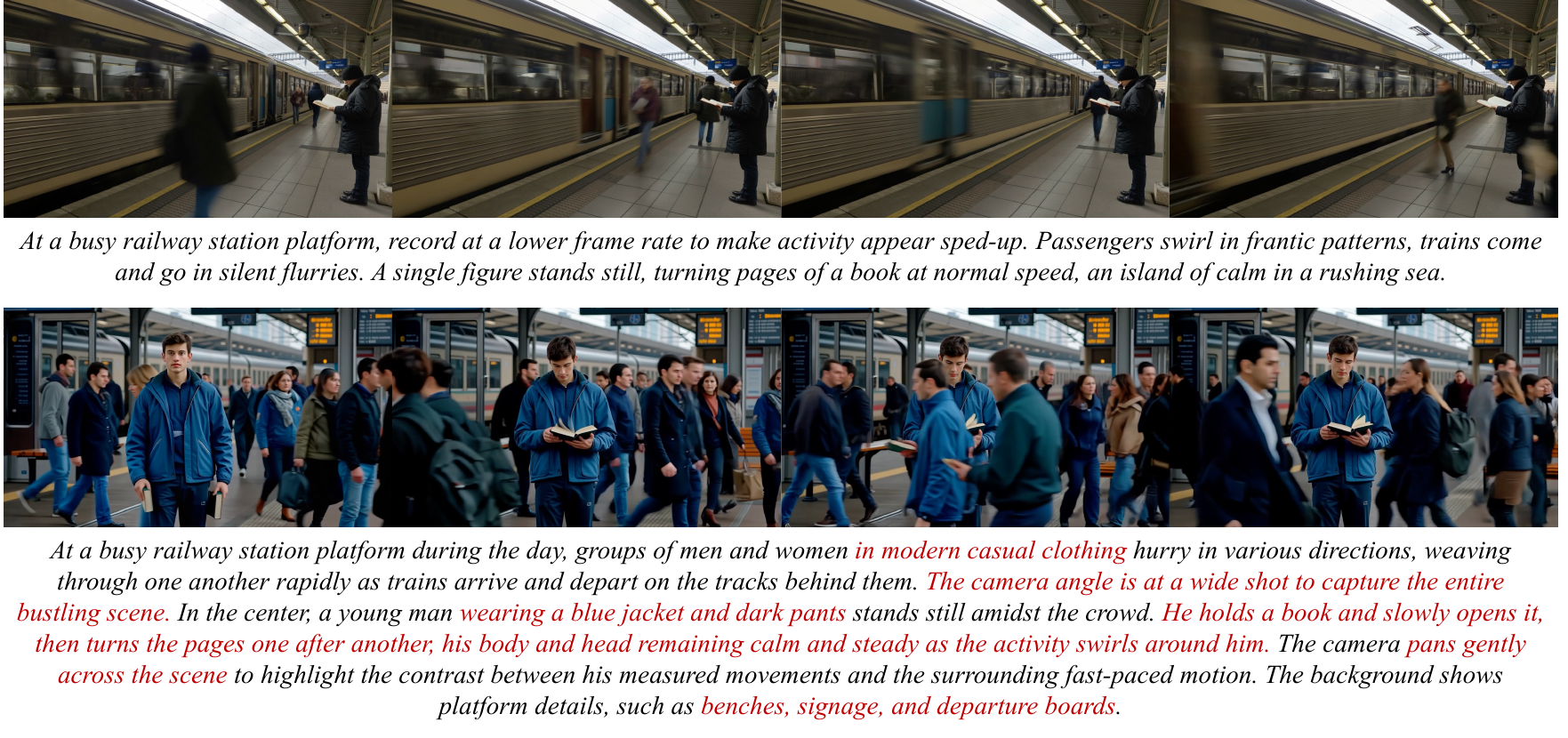}
    \caption{
        \textbf{Top:} 720p T2V results generated with the original prompt. \textbf{Bottom:} 720p T2V results generated with the rewritten prompt.
    }
    \label{fig:rewrite_case}
\end{figure}

The overall strategy involves enriching the prompts with additional details related to camera perspective, appearance, background, and actions. When the action duration is too short, we supplement the prompt with additional action descriptions. Furthermore, we provide several examples of training captions to guide the rewriting model, ensuring that the rewritten prompts are as closely aligned as possible with the distribution of training captions. Fig.~\ref{fig:rewrite_case} shows the generated videos before and after rewriting. We observe that the generation results after prompt rewriting exhibit clear advantages in terms of visual richness and aesthetic quality.

For the two different generation durations, 5 seconds and 10 seconds, we designed two distinct system prompts. The rewrite for the 10-second duration, compared to that for 5 seconds, conditionally incorporates more action descriptions and results in a longer overall prompt. 

%% file: content/5_infra.tex
\section{Infrastructure}

\textbf{Hybrid sharded mode of FSDP. }
Since Fully Sharded Data Parallel (FSDP) is recognized for its simplicity and flexibility in large-scale distributed training, we employ the hybrid sharded mode of FSDP~\citep{zhao2023pytorch} as our primary distributed training solution. Considering that communication bandwidth scales inversely with the total number of GPUs, we adopt a hybrid sharding strategy with inner shard sizes of 64 or 128 to balance per-GPU model memory consumption against communication overhead.

\textbf{Torch compile. }
To maximize performance gains with minimal engineer effort, we leveraged PyTorch's \verb|torch.compile| to compile individual transformer layers into fullgraphs, which enables PyTorch's inductor to automatically fuse underlying CUDA operators together during compilation. 

\textbf{Ulysses sequence parallelism. }
When processing video samples at 720p or 1080p resolution with durations of 8–10 seconds, the input sequence length can reach several hundred thousand tokens, placing considerable demands on GPU memory. To address this, we adopt the Ulysses~\citep{jacobs2023deepspeed} sequence parallelism approach, which distributes the computation of a single input sample along the sequence dimension across sequence-parallel process groups. This strategy significantly alleviates memory pressure and enables support for longer sequences within the constraints of available hardware memory.

\textbf{Bucket Dataloader. }
To support training on videos of arbitrary lengths, we adopted a bucket dataloader to ensure that the token lengths of videos in each sampled batch are identical. 
Given the prevalence of short-duration videos in the training dataset and their low training efficiency, we combined the Sequence Packing algorithm to enhance the training efficiency of short videos. 
Specifically, we used SPFHP~\citep{krell2021efficient} to pack data based on token length, thereby reducing the number of short videos in the training data. 
We then bucketed the data according to the computational load of each sequence and loaded it using the bucket dataloader.

\textbf{Selective Activation Checkpointing. }
To achieve optimal computational performance within the constraints of available hardware memory, we employ torch Selective Activation Checkpointing (SAC) to enable fine-grained selection of operators for recomputation. In our SAC policy, operators with lower recomputation costs, such as element-wise addition and multiplication, are selectively recomputed during the backward pass, while activations of more computationally expensive operators, such as attention and GEMM operations, are retained in memory whenever feasible. This approach effectively balances memory savings and computational overhead.

\textbf{Activation offloading. } 
We further employ an activation offloading strategy to reduce activation memory usage. In this approach, activations produced on the GPU during the forward pass are copied to CPU memory, and subsequently prefetched back to GPU memory several steps ahead of their consumption during the backward pass. The offloading and prefetching processes are executed in a dedicated CUDA stream that overlaps with the main computation stream, resulting in negligible overhead. The memory savings achieved through this strategy enable us to retain a greater number of operators in memory under the SAC policy during training.

\begin{figure}[!t]
    \centering
    \includegraphics[width=1.0\textwidth]{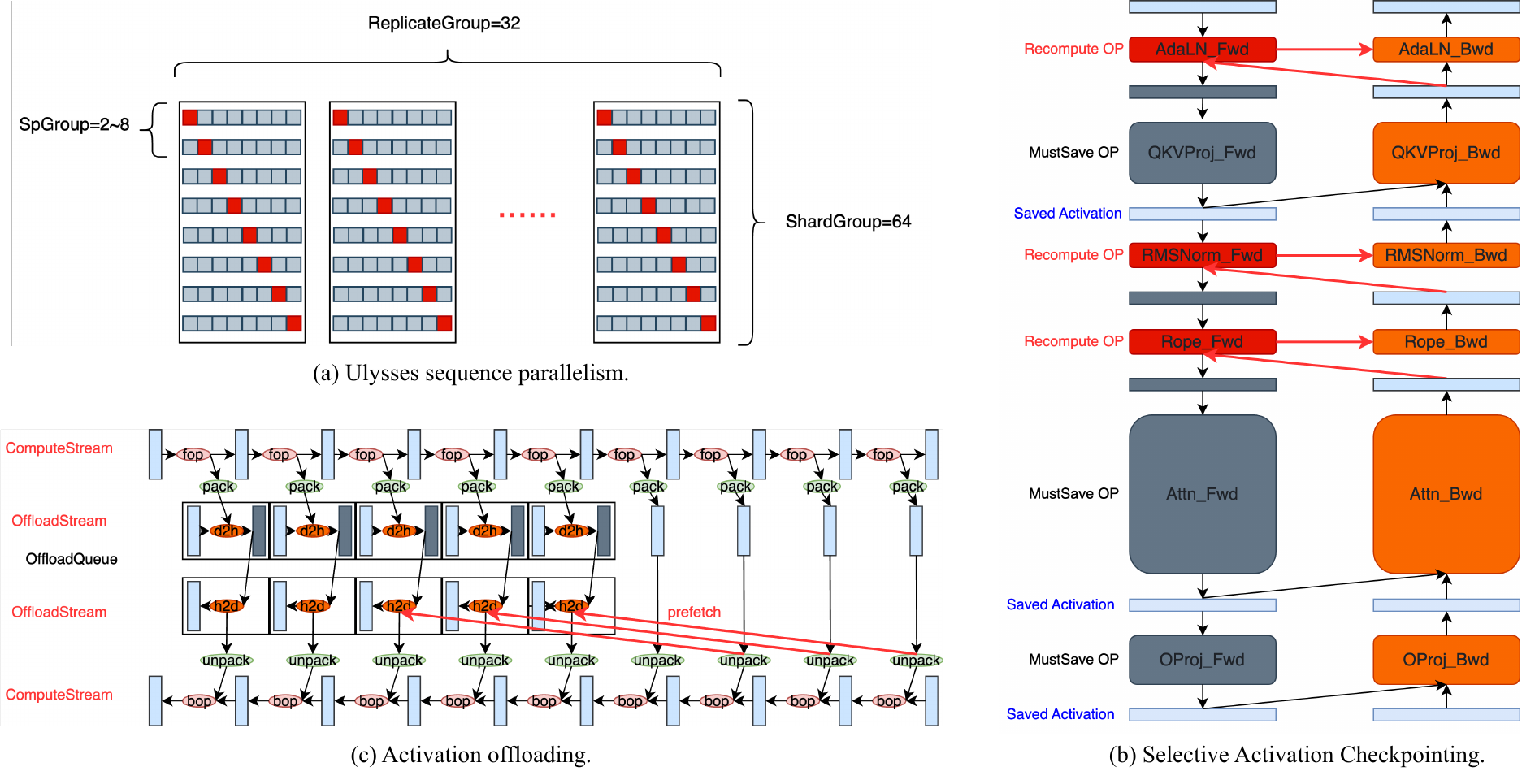}
    \caption{
        Illustrative diagrams of Ulysses sequence parallelism, Selective Activation Checkpointing, and Activation Offloading. In the Activation Offloading diagram, ``fop'' denotes a forward operation, ``bop'' denotes a backward operation, ``h2d'' refers to a host-to-device memory copy, and ``d2h'' refers to a device-to-host memory copy.
    }
    \label{fig:sp}
\end{figure}

\textbf{MFU }
stands for Model FLOPs Utilization. It is a metric that measures the proportion of the theoretical peak floating-point operations per second (FLOPs) of a hardware accelerator (such as a GPU) that is actually utilized by a model during training or inference. Finally, based on the aforementioned optimizations, the MFU achieved during training at different resolutions is summarized in Tab.~\ref{tab:MFU}.


\begin{table}[h]
\centering
\begin{tabular}{lcccc}
\toprule
\textbf{Stages} & \makecell{192p  T2V / T2I} & \makecell{480p  T2V+I2V / T2I} & \makecell{720p  T2V+I2V / T2I} & \makecell{1080p  (T+V)2V / T2I} \\
\midrule
\textbf{MFU}    & 0.32                   & 0.36                      & 0.39                       & 0.40                        \\
\bottomrule
\end{tabular}
\caption{MFU at different training stages.}
\label{tab:MFU}
\end{table}

%% file: content/6_bench.tex
\section{Benchmark Evaluation}

\subsection{Artificial Analysis Arena}
\begin{figure}[htbp]
    \centering
    \includegraphics[width=1.0\textwidth]{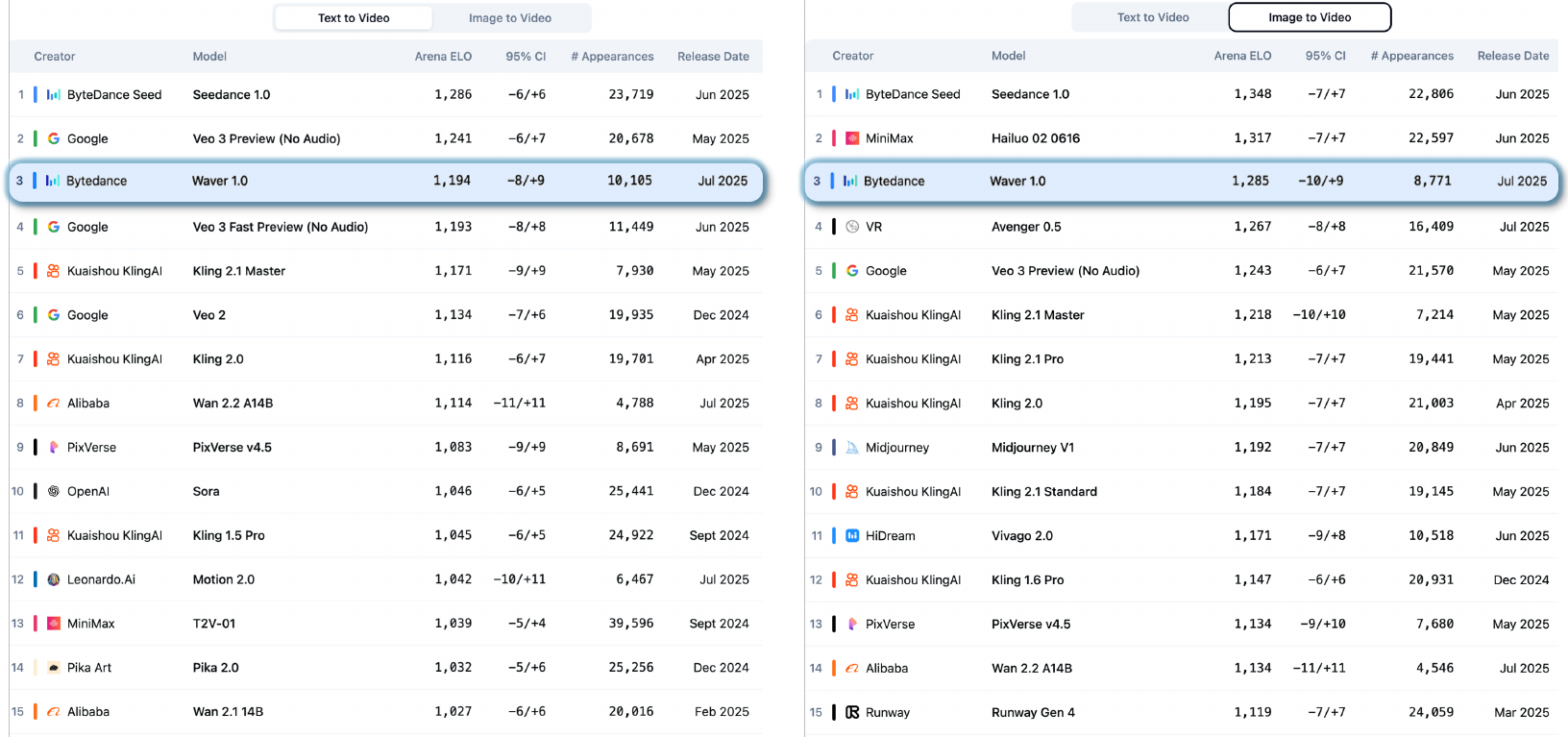}
    \caption{
        On the left and right are the official T2V and I2V ranking results from Artificial Analysis (as of 2025-08-05 12:00 (GMT+8)), with \method ranking third on both lists. 
    }
    \label{fig:arena}
\end{figure}

Artificial Analysis is a popular platform for benchmarking image and video generation models. It allows the public to compare results from different generative models in an open setting. Based on a large number of user comparisons, the platform uses Elo scores to rank models according to user preferences. The ranking results are sourced from the website of the official Artificial Analysis text-to-video (T2V)\footnote{https://artificialanalysis.ai/text-to-video/arena?tab=leaderboard\&input=text} and image-to-video (I2V)\footnote{https://artificialanalysis.ai/text-to-video/arena?tab=leaderboard\&input=image} leaderboard, with data as of 2025-07-22 10:00 (GMT+8). \method ranks third in both the T2V and I2V tracks, as is shown in Fig.~\ref{fig:arena}. 

\subsection{Comprehensive Evaluation}

\paragraph{Evaluation Metrics.}
We define a comprehensive set of metrics for our human evaluation, organized into three main categories.
\textbf{1) Motion Quality.}
This category assesses the physical realism and coherence of movement. This includes: \textit{Action}, which evaluates if the motion is natural and smooth, conforming to the inherent laws of movement; \textit{Interaction}, which checks if interactions (between subjects, subject-object, or between objects) adhere to physical laws; and \textit{Distortion}, which determines if the subject exhibits artifacts like noise, frame loss, or blurring, and ensures the subject remains consistent throughout.
\textbf{2) Visual Quality.}
This category evaluates the aesthetic and technical quality of the image itself. Key aspects are: \textit{Image Quality}, covering texture, lighting, and the presence of visual artifacts such as flickering or overexposure; \textit{Color}, for the appropriateness and appeal of the color scheme; \textit{Clarity}, measuring image sharpness; \textit{Realism}, determining if the visual output is sufficiently realistic (or stylistically consistent for non-realistic styles); and \textit{Aesthetics}, for the overall visual appeal.
\textbf{3) Prompt Following.}
This category measures how faithfully the generated video adheres to the user's text prompt. This is broken down into: \textit{Subject}, ensuring the specified subject(s) (e.g., humans, animals, objects) are present; \textit{Subject Description}, for the accuracy of attributes like number and color; \textit{Action Accuracy}, checking the correctness of actions, temporal sequence, and direction; \textit{Action Magnitude}, assessing the amplitude and speed of actions; and \textit{Camera Movement}, evaluating camera operations like panning, zooming, and tilting.

\begin{figure}[!t]
    \centering
    \includegraphics[width=1.0\textwidth]{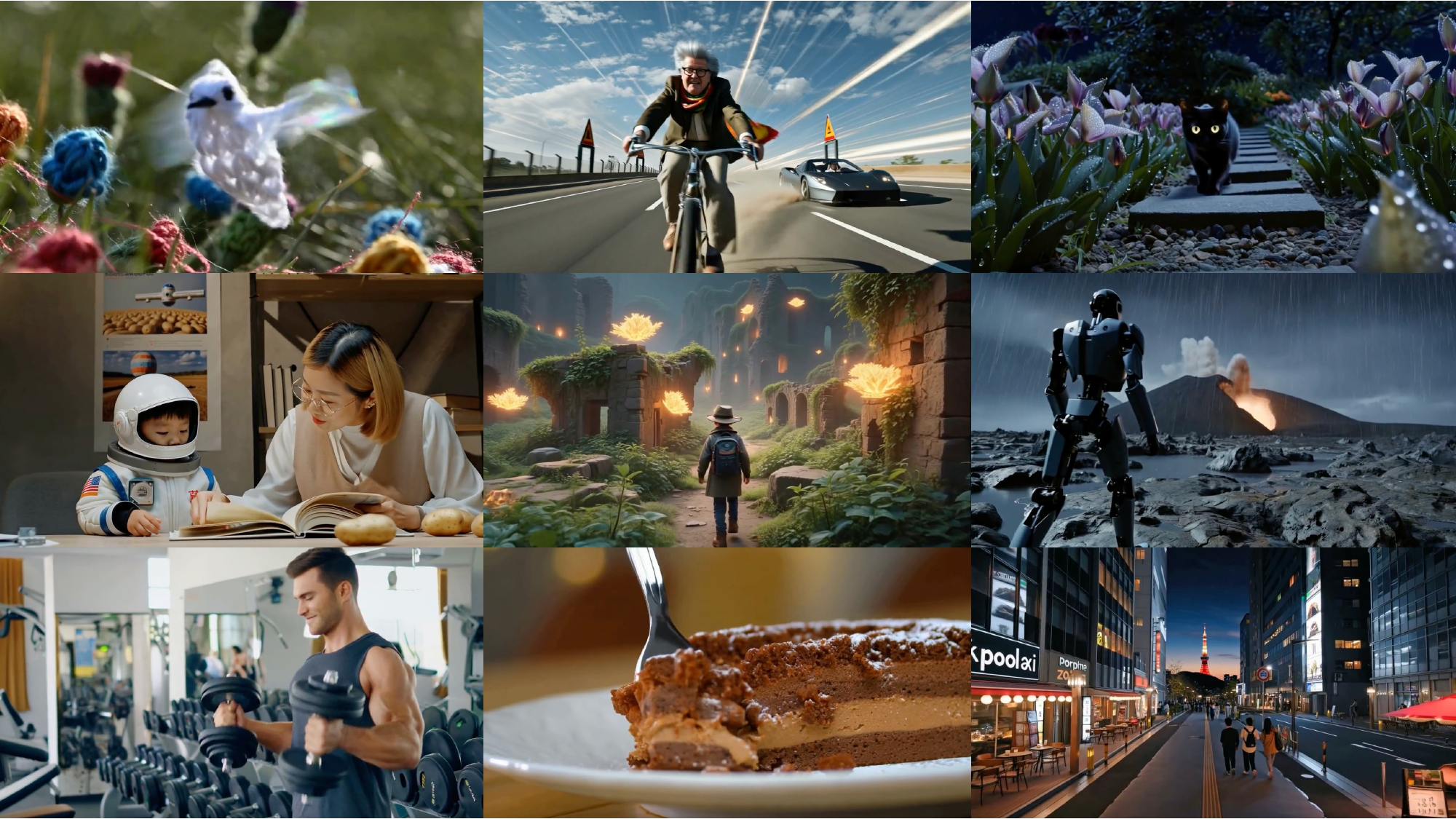}
    \caption{
        Some video examples of the Waver-Bench 1.0 generated by \method. 
    }
    \label{fig:common_show}
\end{figure}

\begin{figure}[!t]
    \centering
    \includegraphics[width=1.0\textwidth]{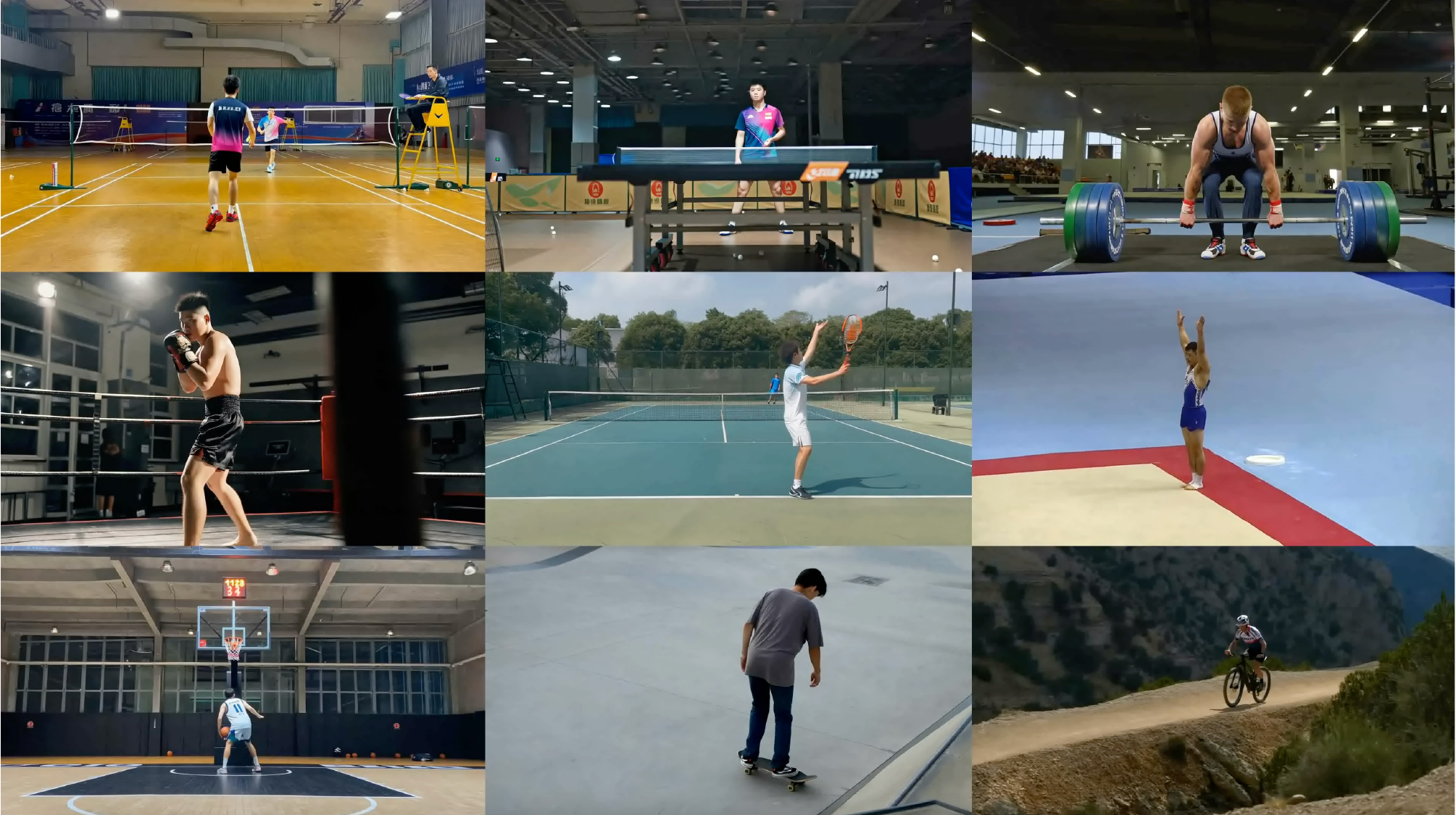}
    \caption{
        Some video examples of the Hermes Motion Testset generated by \method. 
    }
    \label{fig:olympic_show}
\end{figure}

\paragraph{Evaluation Benchmarks.}
To apply these metrics comprehensively and explore the upper limits of our model's motion generation, we introduce two specialized benchmarks. The first, \textbf{Waver-Bench 1.0}, is a broad, general-purpose benchmark consisting of 304 samples that cover a wide range of scenarios, including sports, daily activities, and surreal scenes. The second, the \textbf{Hermes Motion Testset}, is specifically designed to challenge motion generation. It comprises 96 prompts focused on 32 distinct types of sports activities, such as tennis, basketball, and gymnastics. Figure~\ref{fig:common_show} and Figure~\ref{fig:olympic_show} present several generated examples for each benchmark.

\paragraph{Human Evaluation.}
On these benchmarks, we conducted extensive human evaluations to assess our model's performance against leading competitors (Veo3, Kling 2.0, and Wan 2.1 14B). In a side-by-side comparison format, human raters were shown videos generated by our model (Waver) and a competitor, then asked to judge which video was superior. The judgment was based on the comprehensive criteria defined above, including motion quality, visual quality, and prompt following. For the Hermes Motion Testset, we placed a particular emphasis on motion-related sub-dimensions, including Action Accuracy and Action Magnitude.  

\begin{figure}[ht]
    \centering
    \begin{subfigure}{\textwidth}
        \centering
        \includegraphics[width=\textwidth]{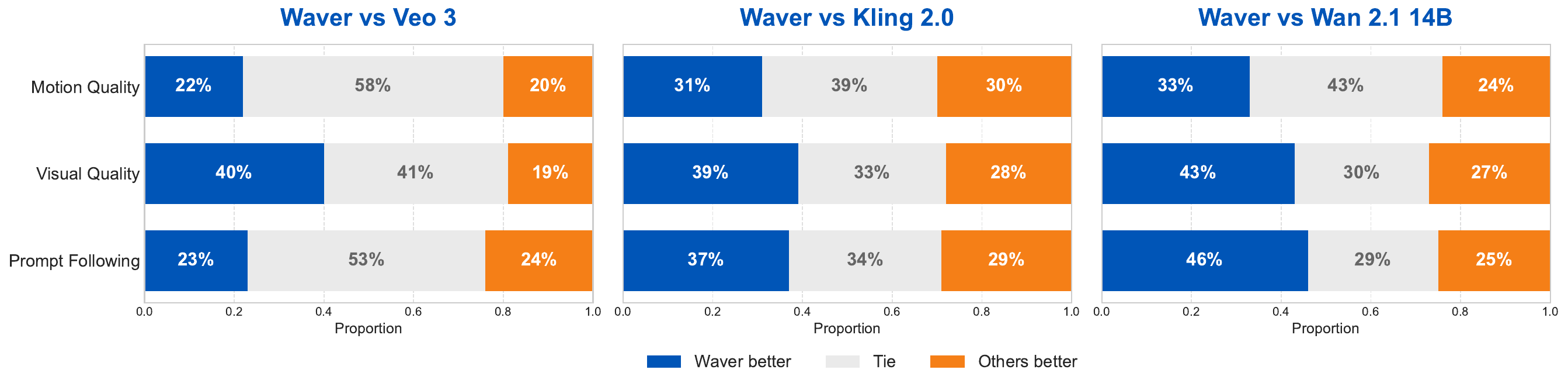}
        \caption{Evaluation on Waver-bench 1.0}
        \label{fig:eval_waver_bench}
    \end{subfigure}
    \vspace{1em} 
    \begin{subfigure}{\textwidth}
        \centering
        \includegraphics[width=\textwidth]{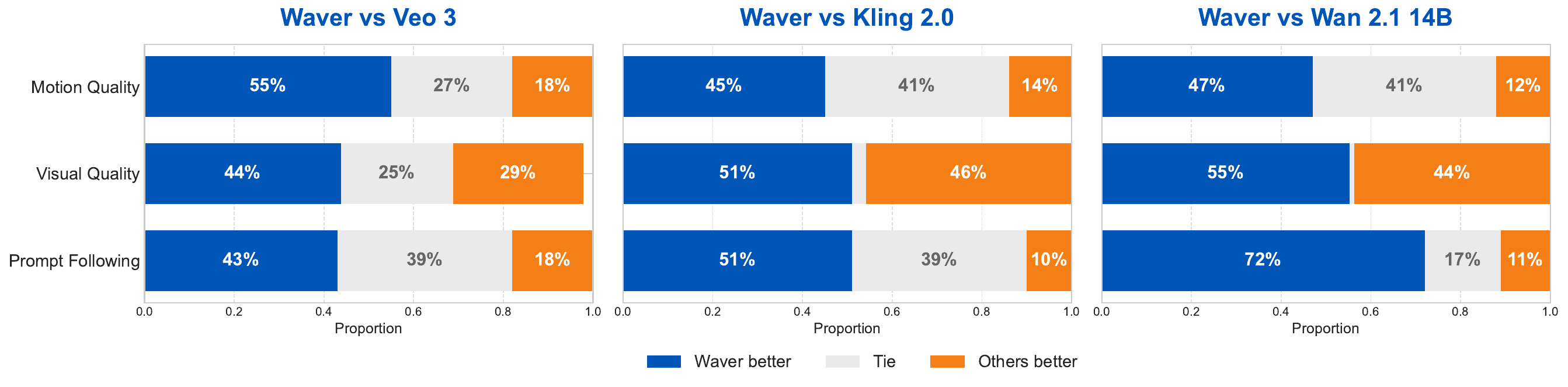}
        \caption{Evaluation on the Hermes Motion Testset}
        \label{fig:eval_hermes_testset}
    \end{subfigure}
    \caption{
        \textbf{Human evaluation results comparing our model (Waver) with leading competitors.}
        Users were presented with side-by-side video comparisons and asked to choose which was better or if they were tied. The stacked bar charts show the proportion of user preference votes across different quality dimensions.
    }
    \label{fig:human_evaluation_combined}
\end{figure}

On the general-purpose \textbf{Waver-bench 1.0} (Figure~\ref{fig:human_evaluation_combined}a), Waver's performance was notably strong. It demonstrated significantly superior motion quality, visual quality, and prompt following when compared to Wan 2.1 14B. Against Kling 2.0, Waver also achieved better visual quality and prompt following, and slightly outperformed it in motion quality. Its performance relative to Veo3 was also competitive, exhibiting higher visual and slightly better motion quality, though its prompt following was marginally weaker.

On the demanding \textbf{Hermes Motion Testset} (Figure~\ref{fig:human_evaluation_combined}b), Waver substantially outperforms all baselines in both motion quality and prompt following. This dominant performance, particularly on a motion-focused benchmark, highlights our model's exceptional capability in generating high-fidelity, dynamic motion that accurately adheres to textual descriptions.

%% file: content/7_discussion.tex
\section{Discussion}

\subsection{DiT Model Sparsity}
\begin{figure}[!t]
    \centering
    \includegraphics[width=0.9\textwidth]{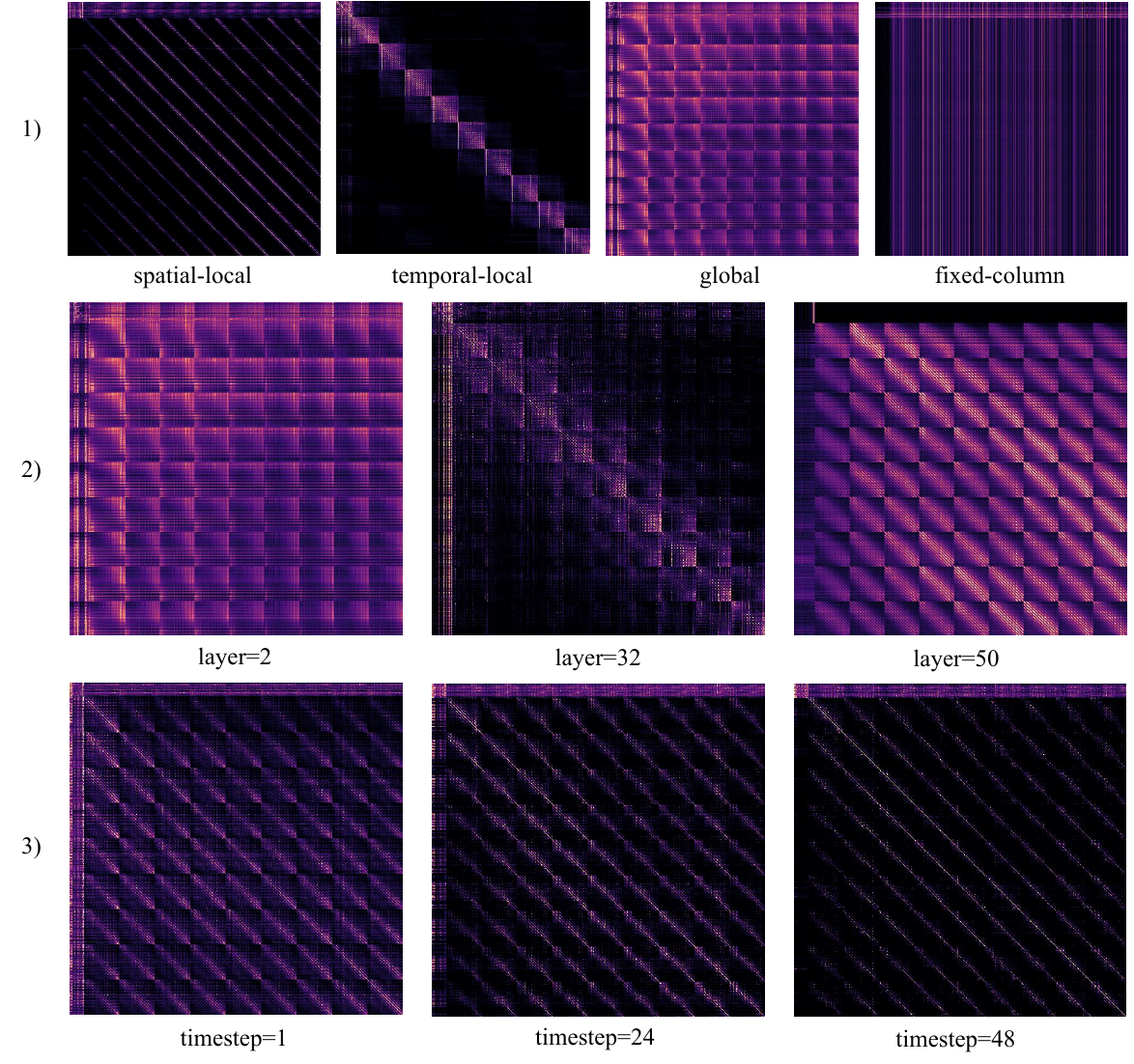}
    \caption{
        An illustration of attention maps is presented, with the query on the vertical axis and the key on the horizontal axis. The visualization reveals three central characteristics: (1) Heterogeneity across attention heads, reflecting functional diversity among heads; (2) Layer-wise sparsity dynamics, showing how attention sparsity shifts across layers; and (3) Timestep-wise sparsity evolution, highlighting changes in attention patterns throughout the denoise process.
    }
    \label{fig:attn-map}
\end{figure}

In DiT, the computational complexity of standard self-attention scales quadratically with the number of input tokens, i.e., $O(N^2)$.
When processing high-resolution, long-duration videos, the number of tokens can easily reach hundreds of thousands, making full self-attention computationally prohibitive and a major performance bottleneck.
For instance, when generating a 5-second 1080p video, attention operations can constitute over 90\% of the total computational cost.
Therefore, reducing computational overhead via sparse attention mechanisms becomes crucial for efficient video generation.

While some approaches~\citep{seawead2025seaweed}—including the one used in our Cascaded Refiner in Sec.\ref{sec:refiner}—leverage window attention to improve efficiency, this method suffers from inherent limitations.
By relying on manually defined, fixed, and non-overlapping windows to constrain the attention range, it fails to capture the natural and content-adaptive patterns typically present in learned attention maps.

To develop sparse attention patterns that better align with video generation dynamics, we conducted a detailed visual analysis of attention maps across different heads, layers, and timesteps.
This analysis revealed three key patterns, as illustrated in Fig.\ref{fig:attn-map}:
1) Heterogeneity Across Attention Heads: Different attention heads exhibit diverse focus patterns, including spatial-local (concentrating on local regions), temporal-local (attending to adjacent frames), global (capturing long-range dependencies), and fixed-column (consistently focusing on specific token columns).
2) Layer-wise Sparsity Dynamics: Attention sparsity varies significantly across layers, following a “dense–sparse–dense” pattern throughout the network.
3) Timestep-wise Sparsity Evolution: The sparsity of attention also evolves over the denoising timesteps, generally increasing as the generation process advances.

These observations suggest that a sparse attention strategy adaptive to the intrinsic structure of attention patterns would outperform fixed-window approaches.
In particular, two directions appear promising:
1) Spatial-Temporal Sliding Window Attention. 
This method offers better alignment with spatio-temporally local attention patterns compared to conventional window attention. 
Although it has been explored in prior work~\citep{xi2025sparse, zhang2025fast}, it remains limited in flexibility and is ineffective at capturing non-local dependencies such as fixed-column patterns.
2) NSA-Based Adaptive Sparse Attention. 
Native Sparse Attention (NSA)~\citep{yuan2025native} provides greater flexibility in designing attention patterns. 
However, existing NSA methods are primarily designed for one-dimensional text data. 
Effectively adapting them to video—a 3D spatio-temporal medium—requires careful redesign to incorporate its structural properties.

Overall, these findings underscore the importance of developing sparse attention mechanisms that are not only computationally efficient but also inherently adaptive to the multi-scale and dynamic nature of video. Such approaches are essential for enabling high-performance, real-time video generation and facilitating wider practical application.

\subsection{VAE Empirical Findings on Video Generation}

Given the critical role of Variational Autoencoders (VAEs) in video generation and their substantial influence on the training of diffusion transformers, our model, while leveraging the Wan 2.1 VAE, still necessitates a dedicated discussion on VAE optimization—given that VAE performance directly underpins video generation quality.

In this section, we elaborate on our VAE training protocols and empirical insights tailored to mitigating key visual quality challenges in video generation, such as perceptible grainy textures, unintended background distortions and grid-like artifacts.

\paragraph{KL Loss is Sensitive to Grainy Textures and Background Distortion}

\begin{figure}[!t]
    \centering
    \includegraphics[width=0.9\textwidth]{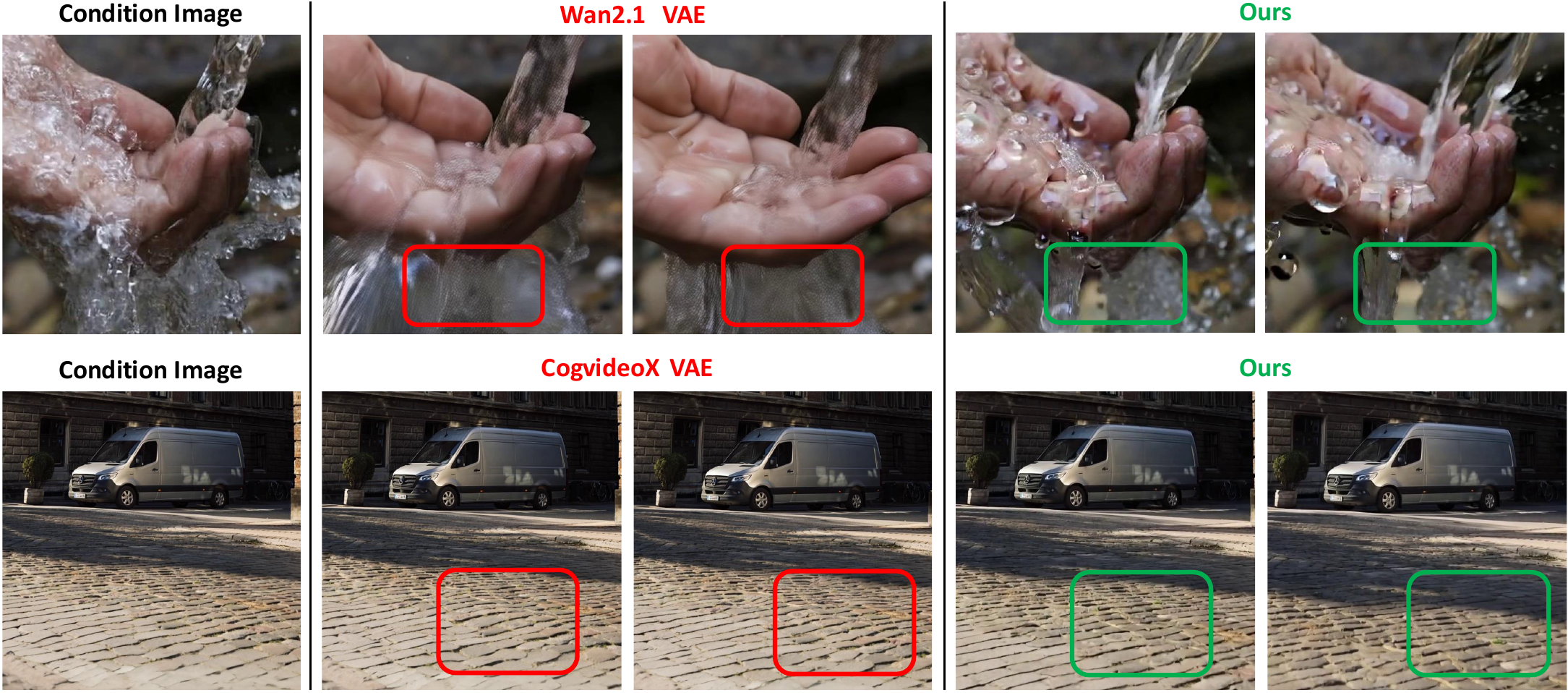}
    \caption{
        Comparison of generated video quality with different VAEs on the I2V task, where the VAEs are trained with different KL loss weights. 
    }
    \label{fig:kl_quality}
\end{figure}

A common practice when training a Video VAE is to use a weighted sum of several loss functions, including L1 reconstruction loss, LPIPS perceptual loss, GAN loss, and KL loss~\citep{cogvideox, wan2025wan, polyak2024moviegencastmedia}.
\begin{equation}
\mathcal{L}_{VAE}=\lambda_1 \mathcal{L}_{1}+\lambda_2 \mathcal{L}_{LPIPS}+\lambda_3 \mathcal{L}_{GAN}+\lambda_4 \mathcal{L}_{KL}. 
\end{equation}

\begin{figure}[htbp]
    \centering
    \includegraphics[width=0.95\textwidth]{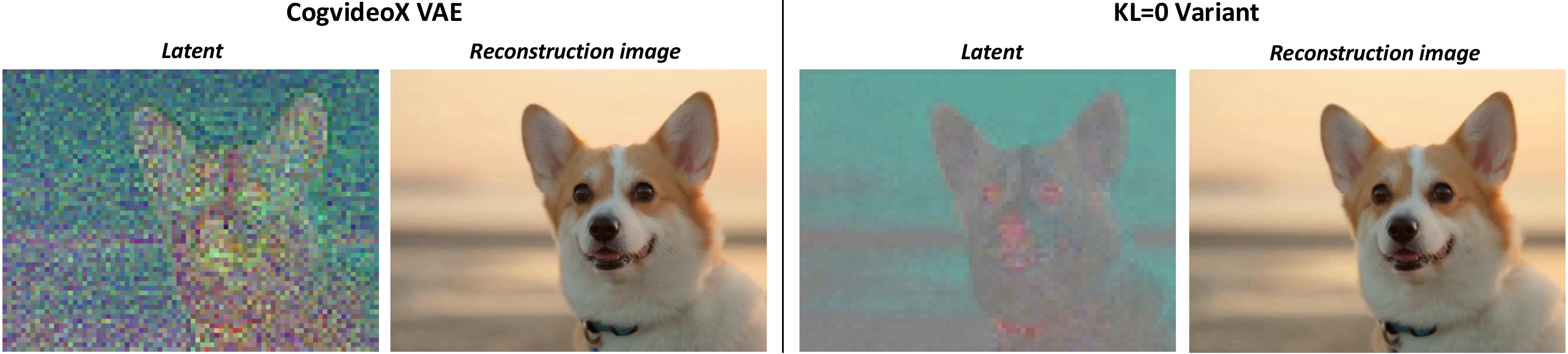}
    \caption{
        Comparison of the latent distributions and reconstructed images between CogVideoX VAE and its KL=0 variant.
    }
    \label{fig:kl0}
\end{figure}

The KL divergence loss in our exp is computed in closed form for the Gaussian case as follows:
\begin{equation}
    \mathcal{L}_{\mathrm{KL}}=\mathrm{KL}\left(\mathcal{N}(\mu,  \sigma^2 I) \parallel  \mathcal{N}(0, I)\right) =\frac{1}{2DTHW} \sum_{d,t,h,w} \left( \sigma_{dthw}^2 + \mu_{dthw}^2 - 1 - \log \sigma_{dthw}^2 \right),
    \label{eq:vae_loss}
\end{equation}
where $D$, $T$, $H$, and $W$ denote the latent dimension, temporal length, height, and width of the latent tokens, respectively. And $\mu_{dthw}$ and $\sigma_{dthw}$ are the mean and standard deviation at each position.
The KL loss serves as a regularization objective, encouraging the learned latent distribution to approximate a standard normal distribution. The weight of the KL loss, denoted as $\lambda_4$ in Eq.~\ref{eq:vae_loss}, controls the strength of the regularization. 

\begin{figure}[htbp]
    \centering
    \includegraphics[width=0.95\textwidth]{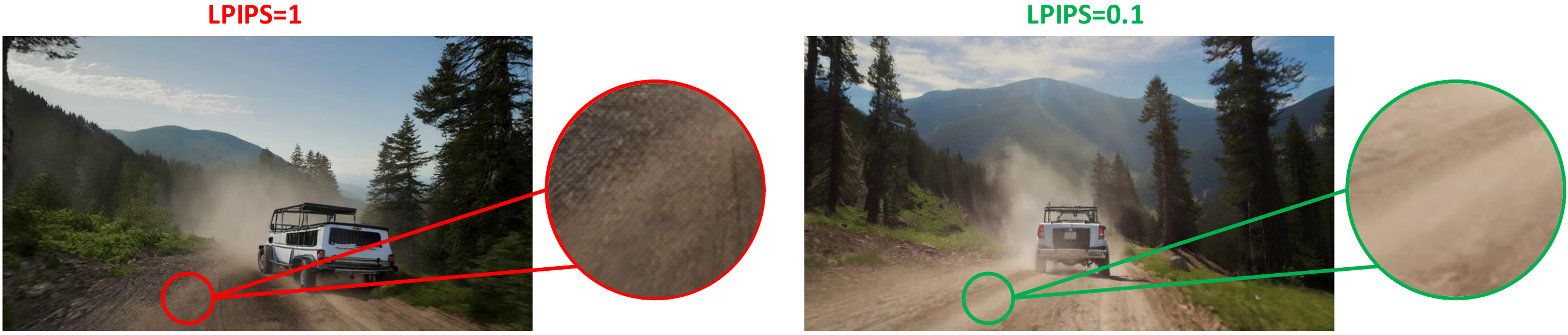}
    \caption{
        Ablation study of video generation results using VAEs trained with different LPIPS losses. 
    }
    \label{fig:lpips}
\end{figure}

We investigate this issue within the I2V task to ensure fair comparison through consistent scene control. By finetuning the Waver I2V model with different VAEs using identical conditional inputs, we compare videos generated by our VAE against those from Wan 2.1 VAE~\citep{wan2025wan} and CogvideoX VAE~\citep{cogvideox} (Fig.~\ref{fig:kl_quality}). Our results indicate that a low KL loss weight (Wan VAE) causes grainy textures due to insufficient decoder robustness, which can lead to mode collapse. Conversely, an excessively high KL weight (CogvideoX VAE) distorts backgrounds. While a larger KL weight enhances decoder robustness through reparameterization-based sampling, finding an optimal balance is crucial.

We visualize the latent distributions of CogvideoX VAE and its KL=0 variant in Fig.~\ref{fig:kl0}. Although both VAEs achieve similar reconstruction, their latent spaces differ significantly. With KL loss weight at 0, the VAE learns a smooth, well-structured latent space. As the KL weight increases, the latent representation collapses toward the prior $\mathcal{N}(0, I)$, becoming noisier and making diffusion model training more challenging.

\begin{table}[!t]
\centering
\scriptsize
\begin{tabularx}{\textwidth}{lX}
\toprule

\textbf{Without spatial relation QA pairs} & In an outdoor courtyard featuring dark tiled flooring, white ceramic planters shaped like toilets, and a backdrop of a wooden-and-metal fence adorned with hanging circular decorations, an older man (around 40 years old) stands near a house entrance. He wears a dark jacket, brown beanie, and blue jeans, holding something small in his hands as he looks toward the scene unfolding nearby. The camera slowly pans right, shifting focus from the man to a woman around 50 years old standing on a concrete path adjacent to the courtyard. She is dressed in a fluffy pink coat, black beanie, and red scarf, \textcolor{red}{with one hand in her pocket and the other gesturing expressively}—first waving, then pointing, and crossing arms—as she speaks. Two dogs accompany her: a brown one with a collar stands calmly beside her, while a white dog moves around her feet. The courtyard’s greenery, including yellow-leafed plants in the foreground and potted shrubs along the fence, frames the scene, with distant buildings and trees visible beyond. The woman continues her animated gestures. \\
\midrule
\textbf{With spatial relation QA pairs}    & In an outdoor courtyard featuring dark tiled flooring, white ceramic planters shaped like toilets, and a wooden-and-metal fence adorned with hanging circular decorations, a man around 40 years old—dressed in a brown beanie, dark jacket, and blue jeans—stands near a house with a covered pergola and red-tiled roof, holding an object in his hands. The camera slowly pans right, shifting focus from the man to reveal a woman around 50 years old on the right side of the frame. She wears a pink fuzzy coat, black beanie, and red scarf, \textcolor{green}{with her left hand in her coat pocket and her right hand making expressive gestures}: first raising it with fingers spread, then pointing toward the camera, and finally crossing her arms. Beside her, two dogs—a brown one with a collar and a white one—stand on the concrete path, occasionally moving slightly. The background includes more potted plants, trees, and distant buildings with green fencing. As the camera continues panning, \textcolor{green}{the man remains in the left background}, still holding his object, \textcolor{green}{while the woman stays centered}, continuing her gestures and smiling. The dogs remain nearby, with the brown dog eventually turning its head toward the camera. The scene maintains a casual, warm atmosphere throughout, with the woman’s movements and the dogs’ presence adding dynamic elements to the static courtyard setting.                       \\
\bottomrule
\end{tabularx}
\caption{Caption model trained without/with spatial relation QA pairs.}
\label{tab:CAP_spatial_relation}
\end{table}

\paragraph{Decrease LPIPS Loss Weight Can Avoid Grid-like Artifacts}
We find that although the LPIPS loss~\citep{zhang2018unreasonable} helps stabilize VAE training, it can also introduce grid-like artifacts in video generation, particularly in detailed regions with high-motion dynamics.
As illustrated in Fig.~\ref{fig:lpips}, the VAE trained with a higher LPIPS loss weight exhibits obvious grid-like artifacts on the ground, whereas the VAE trained with an appropriately reduced LPIPS loss produces a smooth surface.
Based on these observations, we set the LPIPS loss weight to $0.1$ in our experiments.

\paragraph{Exploring Possible Future Directions for Video VAEs}
(1) \textit{Higher Compression Ratios.} Video VAEs with higher compression ratios are vital for efficient DiT training, as they reduce token length and enable longer video generation. The primary challenge is to advance this compression without degrading reconstruction and generation quality. (2) \textit{Fusion with Multimodality.} Future Video VAEs should evolve from visual compressors into multimodal encoders. A key direction is to develop architectures that jointly embed video, audio, and text into a unified latent space, enabling more semantically rich and controllable multimodal generation.

\begin{figure}[!t]
    \centering
    \includegraphics[width=0.95\textwidth]{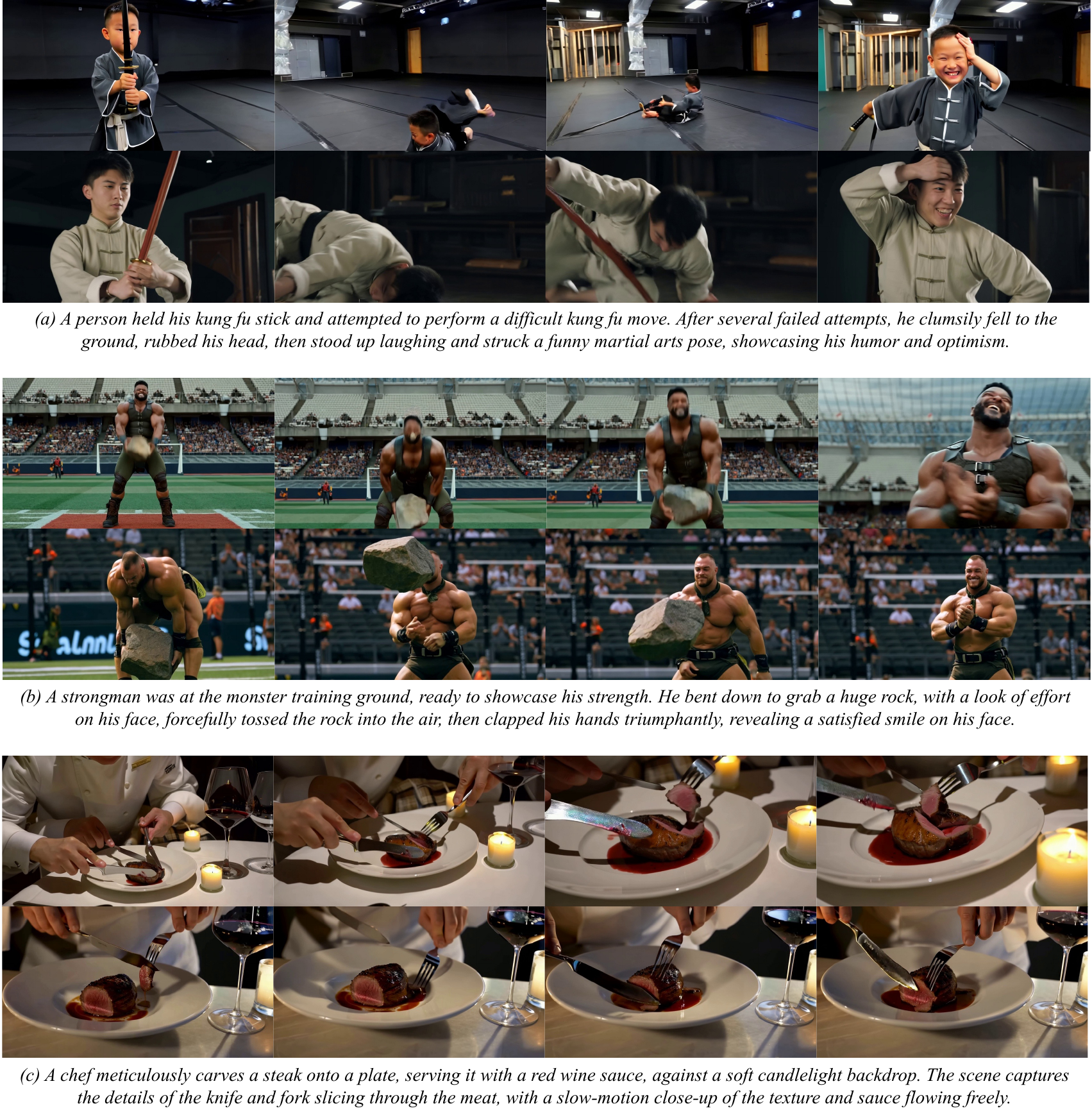}
    \caption{
        \textbf{Top:} The results of the T2V model trained using the Tarsier2 caption model. \textbf{Bottom:} The outcomes of the T2V model trained with our proposed caption model. In case (a), there is a noticeable distortion in the legs in the top row, which is absent in the bottom row. In case (b), the strongman repeatedly lifts and lowers the stone in the top row, whereas in the bottom row, the actions strictly follow the given instructions. In case (c), the character is depicted with three hands, and the knife held in one hand is incorrectly rendered as a fork in the top row. In contrast, the bottom row accurately shows the chef holding a knife in one hand and a fork in the other.
    }
    \label{fig:caption_case}
\end{figure}

\subsection{Caption Impact on Video Generation}
\paragraph{Detailed Caption is Indeed Important for Precise Instruction Following}
The use of detailed caption models facilitates a more precise alignment between video content and its corresponding captions, enabling more effective training of Text-to-Video (T2V) models, which contributes to a reduction in motion distortion in the pretraining phase and improves the model’s ability to generate visuals that are more consistent with real world scenes. Furthermore, since detailed captions describe actions with greater specificity and adhere strictly to temporal order, T2V models exhibit enhanced instruction-following capabilities with respect to action sequences. Fig.~\ref{fig:caption_case} illustrates the advantages of our T2V model trained with more detailed captions compared to that trained with Tarsier2~\citep{yuan2025tarsier2} captions.

\paragraph{Enhancing Spatial Relation Representation is Helpful}
Due to the lack of explicit indication regarding whether spatial relations are described from the viewer's perspective or from the perspective of the subjects within the video in publicly available Multimodal Large Language Model (MLLM) pretraining datasets, MLLMs often make errors when generating captions that reference spatial relations such as ``left'' or ``right''. Specifically, models may incorrectly specify the spatial relation, omit the relevant perspective, or completely fail to describe it. This ambiguity negatively impacts the alignment between the video content and the generated caption.

To address this issue in subsequent optimization of the caption model, we place particular emphasis on the accurate representation of spatial relations. We construct a comprehensive set of question-answer (QA) pairs focused on spatial relations, with each question explicitly requiring temporal/sequential localization, subject/object localization, and action localization. This design ensures that each question refers to a unique event or moment within the video. The cases of caption model trained without/with spatial relation QA pairs are presented in Tab.~\ref{tab:CAP_spatial_relation}.

%% file: content/8_conclusion.tex
\section{Conclusion and Limitation}
\label{sec:conclusion}
\paragraph{Conclusion.}
We introduce \method, a rectified flow Transformer framework that unifies multiple video generation tasks and achieves high-performance generation results. Through a robust data curation pipeline and detailed optimization strategies, \method demonstrates superior performance across public and internal benchmarks, especially in motion quality. We hope that the insights and practical recipes provided in this work will support the community in further advancing the state of video generation technology. 

\paragraph{Limitation.}
There also exhibits certain limitations. For instance, in high-motion scenarios, human body details such as hands and legs are prone to distortion. Additionally, the generated videos sometimes lack rich visual details, resulting in limited expressiveness. In future work, we will employ reinforcement learning (RL) techniques to mitigate these distortions and enhance visual details, thereby further improving the generation quality.

%% file: content/9_authors.tex
\section{Contributions and Acknowledgments}
\label{contributions}
\textbf{Core Contributors: }

Yifu Zhang, Hao Yang, Yuqi Zhang, Yifei Hu, Fengda Zhu, Chuang Lin, Xiaofeng Mei, Yi Jiang, Bingyue Peng, Zehuan Yuan

\textbf{Contributors: }

Ruibiao Lu, Prasanna Raghav, Yi Fu, Zeyu Zhang, Hui Wu, Xing Wang, Chongxi Wang, Xibin Wu,  Hongxiang Hao, Heng Zhang, Yanghua Peng

\textbf{Acknowledgments: }

Fangzhou Ai, Jinlai Liu, Xugang Ye, Xiaoran Xu, Bobo Zeng, Aaron Shen, Mao Hu, Xiaoxiao Qin, Tingxuan Li, Wanxing Wang, Puke Zhang, Yufei Wu, Ruoyu Guo, Ge Bai, Xin Chen, Dongyang Wang, Tiger Li, Shoufa Chen, Chongjian Ge, Shilong Zhang, Shukai Wang, Dingyuan Xu, Chetan Velivela, Kairong Sun, Kaihua Jiang, Junru Zheng

\clearpage

%% file: main.bbl
\begin{thebibliography}{45}
\providecommand{\natexlab}[1]{#1}
\providecommand{\url}[1]{\texttt{#1}}
\expandafter\ifx\csname urlstyle\endcsname\relax
  \providecommand{\doi}[1]{doi: #1}\else
  \providecommand{\doi}{doi: \begingroup \urlstyle{rm}\Url}\fi

\bibitem[Bai et~al.(2025)Bai, Chen, Liu, Wang, Ge, Song, Dang, Wang, Wang, Tang, et~al.]{bai2025qwen2}
Shuai Bai, Keqin Chen, Xuejing Liu, Jialin Wang, Wenbin Ge, Sibo Song, Kai Dang, Peng Wang, Shijie Wang, Jun Tang, et~al.
\newblock Qwen2. 5-vl technical report.
\newblock \emph{arXiv preprint arXiv:2502.13923}, 2025.

\bibitem[Blattmann et~al.(2023)Blattmann, Dockhorn, Kulal, Mendelevitch, Kilian, Lorenz, Levi, English, Voleti, Letts, et~al.]{blattmann2023stable}
Andreas Blattmann, Tim Dockhorn, Sumith Kulal, Daniel Mendelevitch, Maciej Kilian, Dominik Lorenz, Yam Levi, Zion English, Vikram Voleti, Adam Letts, et~al.
\newblock Stable video diffusion: Scaling latent video diffusion models to large datasets.
\newblock \emph{arXiv preprint arXiv:2311.15127}, 2023.

\bibitem[Chen et~al.(2025)Chen, Ge, Zhang, Zhang, Zhu, Yang, Hao, Wu, Lai, Hu, et~al.]{chen2025goku}
Shoufa Chen, Chongjian Ge, Yuqi Zhang, Yida Zhang, Fengda Zhu, Hao Yang, Hongxiang Hao, Hui Wu, Zhichao Lai, Yifei Hu, et~al.
\newblock Goku: Flow based video generative foundation models.
\newblock In \emph{Proceedings of the Computer Vision and Pattern Recognition Conference}, pp.\  23516--23527, 2025.

\bibitem[DeepMind(2025.05)]{veo32025}
Google DeepMind.
\newblock Veo 3.
\newblock \emph{https://deepmind.google/technologies/veo/veo-3/}, 2025.05.

\bibitem[Developers()]{pyscenedetect}
PySceneDetect Developers.
\newblock Pyscenedetect.
\newblock URL \url{https://www.scenedetect.com}.

\bibitem[Dosovitskiy et~al.(2020)Dosovitskiy, Beyer, Kolesnikov, Weissenborn, Zhai, Unterthiner, Dehghani, Minderer, Heigold, Gelly, et~al.]{dosovitskiy2020image}
Alexey Dosovitskiy, Lucas Beyer, Alexander Kolesnikov, Dirk Weissenborn, Xiaohua Zhai, Thomas Unterthiner, Mostafa Dehghani, Matthias Minderer, Georg Heigold, Sylvain Gelly, et~al.
\newblock An image is worth 16x16 words: Transformers for image recognition at scale.
\newblock \emph{arXiv preprint arXiv:2010.11929}, 2020.

\bibitem[Esser et~al.(2024)Esser, Kulal, Blattmann, Entezari, M{\"u}ller, Saini, Levi, Lorenz, Sauer, Boesel, et~al.]{esser2024sd3}
Patrick Esser, Sumith Kulal, Andreas Blattmann, Rahim Entezari, Jonas M{\"u}ller, Harry Saini, Yam Levi, Dominik Lorenz, Axel Sauer, Frederic Boesel, et~al.
\newblock Scaling rectified flow transformers for high-resolution image synthesis.
\newblock In \emph{Int. Conf. Mach. Learn.}, 2024.

\bibitem[Gao et~al.(2025)Gao, Guo, Hoang, Huang, Jiang, Kong, Li, Li, Li, Li, et~al.]{gao2025seedance}
Yu~Gao, Haoyuan Guo, Tuyen Hoang, Weilin Huang, Lu~Jiang, Fangyuan Kong, Huixia Li, Jiashi Li, Liang Li, Xiaojie Li, et~al.
\newblock Seedance 1.0: Exploring the boundaries of video generation models.
\newblock \emph{arXiv preprint arXiv:2506.09113}, 2025.

\bibitem[Girdhar et~al.(2023)Girdhar, Singh, Brown, Duval, Azadi, Rambhatla, Shah, Yin, Parikh, and Misra]{girdhar2023emu}
Rohit Girdhar, Mannat Singh, Andrew Brown, Quentin Duval, Samaneh Azadi, Sai~Saketh Rambhatla, Akbar Shah, Xi~Yin, Devi Parikh, and Ishan Misra.
\newblock Emu video: Factorizing text-to-video generation by explicit image conditioning.
\newblock \emph{arXiv preprint arXiv:2311.10709}, 2023.

\bibitem[Jacobs et~al.(2023)Jacobs, Tanaka, Zhang, Zhang, Song, Rajbhandari, and He]{jacobs2023deepspeed}
Sam~Ade Jacobs, Masahiro Tanaka, Chengming Zhang, Minjia Zhang, Shuaiwen~Leon Song, Samyam Rajbhandari, and Yuxiong He.
\newblock Deepspeed ulysses: System optimizations for enabling training of extreme long sequence transformer models.
\newblock \emph{arXiv preprint arXiv:2309.14509}, 2023.

\bibitem[Kong et~al.(2024)Kong, Tian, Zhang, Min, Dai, Zhou, Xiong, Li, Wu, Zhang, et~al.]{kong2024hunyuanvideo}
Weijie Kong, Qi~Tian, Zijian Zhang, Rox Min, Zuozhuo Dai, Jin Zhou, Jiangfeng Xiong, Xin Li, Bo~Wu, Jianwei Zhang, et~al.
\newblock Hunyuanvideo: A systematic framework for large video generative models.
\newblock \emph{arXiv preprint arXiv:2412.03603}, 2024.

\bibitem[Krell et~al.(2021)Krell, Kosec, Perez, and Fitzgibbon]{krell2021efficient}
Mario~Michael Krell, Matej Kosec, Sergio~P Perez, and Andrew Fitzgibbon.
\newblock Efficient sequence packing without cross-contamination: Accelerating large language models without impacting performance.
\newblock \emph{arXiv preprint arXiv:2107.02027}, 2021.

\bibitem[Kuaishou(2024.06)]{kuaishou2024kling}
Kuaishou.
\newblock Kling ai.
\newblock \emph{https://klingai.kuaishou.com/}, 2024.06.

\bibitem[Li et~al.(2025)Li, Ma, Yan, Zhang, Liu, Lu, Xu, Chen, Wang, Zhan, et~al.]{li2025model}
Yunshui Li, Yiyuan Ma, Shen Yan, Chaoyi Zhang, Jing Liu, Jianqiao Lu, Ziwen Xu, Mengzhao Chen, Minrui Wang, Shiyi Zhan, et~al.
\newblock Model merging in pre-training of large language models.
\newblock \emph{arXiv preprint arXiv:2505.12082}, 2025.

\bibitem[Liu et~al.(2025)Liu, Ren, Liu, Xie, Zheng, Zhang, Lu, and Yang]{liu2025dynamic}
Peng Liu, Xiaoming Ren, Fengkai Liu, Qingsong Xie, Quanlong Zheng, Yanhao Zhang, Haonan Lu, and Yujiu Yang.
\newblock Dynamic-i2v: Exploring image-to-video generaion models via multimodal llm.
\newblock \emph{arXiv preprint arXiv:2505.19901}, 2025.

\bibitem[Ma et~al.(2025)Ma, Huang, Yan, Chen, Duan, Yin, Wan, Ming, Song, Chen, et~al.]{ma2025step}
Guoqing Ma, Haoyang Huang, Kun Yan, Liangyu Chen, Nan Duan, Shengming Yin, Changyi Wan, Ranchen Ming, Xiaoniu Song, Xing Chen, et~al.
\newblock Step-video-t2v technical report: The practice, challenges, and future of video foundation model.
\newblock \emph{arXiv preprint arXiv:2502.10248}, 2025.

\bibitem[MiniMax(2024.09)]{minimax2024hailuo}
MiniMax.
\newblock Hailuo ai.
\newblock \emph{https://hailuoai.com/video}, 2024.09.

\bibitem[OpenAI(2024)]{openaisora2024}
OpenAI.
\newblock Video generation models as world simulators, 2024.
\newblock URL \url{https://openai.com/index/video-generation-models-as-world-simulators/}.

\bibitem[Oquab et~al.(2023)Oquab, Darcet, Moutakanni, Vo, Szafraniec, Khalidov, Fernandez, Haziza, Massa, El-Nouby, et~al.]{oquab2023dinov2}
Maxime Oquab, Timoth{\'e}e Darcet, Th{\'e}o Moutakanni, Huy Vo, Marc Szafraniec, Vasil Khalidov, Pierre Fernandez, Daniel Haziza, Francisco Massa, Alaaeldin El-Nouby, et~al.
\newblock Dinov2: Learning robust visual features without supervision.
\newblock \emph{arXiv preprint arXiv:2304.07193}, 2023.

\bibitem[Pal et~al.(2024)Pal, Karkhanis, Dooley, Roberts, Naidu, and White]{pal2024dpop}
Arka Pal, Deep Karkhanis, Samuel Dooley, Manley Roberts, Siddartha Naidu, and Colin White.
\newblock Smaug: Fixing failure modes of preference optimisation with dpo-positive.
\newblock \emph{arXiv preprint arXiv:2402.13228}, 2024.

\bibitem[Peebles \& Xie(2023)Peebles and Xie]{dit}
William Peebles and Saining Xie.
\newblock Scalable diffusion models with transformers.
\newblock In \emph{Int. Conf. Comput. Vis.}, pp.\  4195--4205, 2023.

\bibitem[Polyak et~al.()Polyak, Zohar, Brown, Tjandra, Sinha, Lee, Vyas, Shi, Ma, Chuang, Yan, Choudhary, Wang, Sethi, Pang, Ma, Misra, Hou, Wang, Jagadeesh, Li, Zhang, Singh, Williamson, Le, Yu, Singh, Zhang, Vajda, Duval, Girdhar, Sumbaly, Rambhatla, Tsai, Azadi, Datta, Chen, Bell, Ramaswamy, Sheynin, Bhattacharya, Motwani, Xu, Li, Hou, Hsu, Yin, Dai, Taigman, Luo, Liu, Wu, Zhao, Kirstain, He, He, Pumarola, Thabet, Sanakoyeu, Mallya, Guo, Araya, Kerr, Wood, Liu, Peng, Vengertsev, Schonfeld, Blanchard, Juefei-Xu, Nord, Liang, Hoffman, Kohler, Fire, Sivakumar, Chen, Yu, Gao, Georgopoulos, Moritz, Sampson, Li, Parmeggiani, Fine, Fowler, Petrovic, and Du]{polyak2024moviegencastmedia}
Adam Polyak, Amit Zohar, Andrew Brown, Andros Tjandra, Animesh Sinha, Ann Lee, Apoorv Vyas, Bowen Shi, Chih-Yao Ma, Ching-Yao Chuang, David Yan, Dhruv Choudhary, Dingkang Wang, Geet Sethi, Guan Pang, Haoyu Ma, Ishan Misra, Ji~Hou, Jialiang Wang, Kiran Jagadeesh, Kunpeng Li, Luxin Zhang, Mannat Singh, Mary Williamson, Matt Le, Matthew Yu, Mitesh~Kumar Singh, Peizhao Zhang, Peter Vajda, Quentin Duval, Rohit Girdhar, Roshan Sumbaly, Sai~Saketh Rambhatla, Sam Tsai, Samaneh Azadi, Samyak Datta, Sanyuan Chen, Sean Bell, Sharadh Ramaswamy, Shelly Sheynin, Siddharth Bhattacharya, Simran Motwani, Tao Xu, Tianhe Li, Tingbo Hou, Wei-Ning Hsu, Xi~Yin, Xiaoliang Dai, Yaniv Taigman, Yaqiao Luo, Yen-Cheng Liu, Yi-Chiao Wu, Yue Zhao, Yuval Kirstain, Zecheng He, Zijian He, Albert Pumarola, Ali Thabet, Artsiom Sanakoyeu, Arun Mallya, Baishan Guo, Boris Araya, Breena Kerr, Carleigh Wood, Ce~Liu, Cen Peng, Dimitry Vengertsev, Edgar Schonfeld, Elliot Blanchard, Felix Juefei-Xu, Fraylie Nord, Jeff Liang, John Hoffman, Jonas
  Kohler, Kaolin Fire, Karthik Sivakumar, Lawrence Chen, Licheng Yu, Luya Gao, Markos Georgopoulos, Rashel Moritz, Sara~K. Sampson, Shikai Li, Simone Parmeggiani, Steve Fine, Tara Fowler, Vladan Petrovic, and Yuming Du.
\newblock Movie gen: A cast of media foundation models.
\newblock \emph{arXiv preprint arXiv:2410.13720}.

\bibitem[QwenTeam(2024)]{qwen2.5}
QwenTeam.
\newblock Qwen2.5: A party of foundation models, September 2024.
\newblock URL \url{https://qwenlm.github.io/blog/qwen2.5/}.

\bibitem[Rafailov et~al.(2024)Rafailov, Sharma, Mitchell, Ermon, Manning, and Finn]{rafailov2024dpo}
Rafael Rafailov, Archit Sharma, Eric Mitchell, Stefano Ermon, Christopher~D. Manning, and Chelsea Finn.
\newblock Direct preference optimization: Your language model is secretly a reward model.
\newblock \emph{arXiv preprint arXiv:2305.18290}, 2024.

\bibitem[Raffel et~al.(2020)Raffel, Shazeer, Roberts, Lee, Narang, Matena, Zhou, Li, and Liu]{t5}
Colin Raffel, Noam Shazeer, Adam Roberts, Katherine Lee, Sharan Narang, Michael Matena, Yanqi Zhou, Wei Li, and Peter~J Liu.
\newblock Exploring the limits of transfer learning with a unified text-to-text transformer.
\newblock \emph{The Journal of Machine Learning Research}, 21\penalty0 (1):\penalty0 5485--5551, 2020.

\bibitem[Sadat et~al.(2024)Sadat, Hilliges, and Weber]{sadat2024eliminating}
Seyedmorteza Sadat, Otmar Hilliges, and Romann~M Weber.
\newblock Eliminating oversaturation and artifacts of high guidance scales in diffusion models.
\newblock In \emph{The Thirteenth International Conference on Learning Representations}, 2024.

\bibitem[Seawead et~al.(2025)Seawead, Yang, Lin, Zhao, Lin, Ma, Guo, Chen, Qi, Wang, et~al.]{seawead2025seaweed}
Team Seawead, Ceyuan Yang, Zhijie Lin, Yang Zhao, Shanchuan Lin, Zhibei Ma, Haoyuan Guo, Hao Chen, Lu~Qi, Sen Wang, et~al.
\newblock Seaweed-7b: Cost-effective training of video generation foundation model.
\newblock \emph{arXiv preprint arXiv:2504.08685}, 2025.

\bibitem[Singer et~al.(2022)Singer, Polyak, Hayes, Yin, An, Zhang, Hu, Yang, Ashual, Gafni, et~al.]{singer2022make}
Uriel Singer, Adam Polyak, Thomas Hayes, Xi~Yin, Jie An, Songyang Zhang, Qiyuan Hu, Harry Yang, Oron Ashual, Oran Gafni, et~al.
\newblock Make-a-video: Text-to-video generation without text-video data.
\newblock \emph{arXiv preprint arXiv:2209.14792}, 2022.

\bibitem[Su et~al.(2024)Su, Ahmed, Lu, Pan, Bo, and Liu]{su2024roformer}
Jianlin Su, Murtadha Ahmed, Yu~Lu, Shengfeng Pan, Wen Bo, and Yunfeng Liu.
\newblock Roformer: Enhanced transformer with rotary position embedding.
\newblock \emph{Neurocomputing}, 568:\penalty0 127063, 2024.

\bibitem[Teed \& Deng(2021)Teed and Deng]{2021RAFT}
Zachary Teed and Jia Deng.
\newblock Raft: Recurrent all-pairs field transforms for optical flow (extended abstract).
\newblock In \emph{Assoc. Adv. Artif. Intell.}, 2021.

\bibitem[Tian et~al.(2025)Tian, Qu, Lu, Wei, Liu, and Cheng]{tian2025extrapolating}
Jie Tian, Xiaoye Qu, Zhenyi Lu, Wei Wei, Sichen Liu, and Yu~Cheng.
\newblock Extrapolating and decoupling image-to-video generation models: Motion modeling is easier than you think.
\newblock In \emph{Proceedings of the Computer Vision and Pattern Recognition Conference}, pp.\  12512--12521, 2025.

\bibitem[Wan et~al.(2025)Wan, Wang, Ai, Wen, Mao, Xie, Chen, Yu, Zhao, Yang, et~al.]{wan2025wan}
Team Wan, Ang Wang, Baole Ai, Bin Wen, Chaojie Mao, Chen-Wei Xie, Di~Chen, Feiwu Yu, Haiming Zhao, Jianxiao Yang, et~al.
\newblock Wan: Open and advanced large-scale video generative models.
\newblock \emph{arXiv preprint arXiv:2503.20314}, 2025.

\bibitem[Xi et~al.(2025)Xi, Yang, Zhao, Xu, Li, Li, Lin, Cai, Zhang, Li, et~al.]{xi2025sparse}
Haocheng Xi, Shuo Yang, Yilong Zhao, Chenfeng Xu, Muyang Li, Xiuyu Li, Yujun Lin, Han Cai, Jintao Zhang, Dacheng Li, et~al.
\newblock Sparse videogen: Accelerating video diffusion transformers with spatial-temporal sparsity.
\newblock \emph{arXiv preprint arXiv:2502.01776}, 2025.

\bibitem[Yang et~al.(2025)Yang, Teng, Zheng, Ding, Huang, Xu, Yang, Hong, Zhang, Feng, Yin, Gu, Zhang, Wang, Cheng, Liu, Xu, Dong, and Tang]{cogvideox}
Zhuoyi Yang, Jiayan Teng, Wendi Zheng, Ming Ding, Shiyu Huang, Jiazheng Xu, Yuanming Yang, Wenyi Hong, Xiaohan Zhang, Guanyu Feng, Da~Yin, Xiaotao Gu, Yuxuan Zhang, Weihan Wang, Yean Cheng, Ting Liu, Bin Xu, Yuxiao Dong, and Jie Tang.
\newblock {{CogVideoX}}: {{Text-to-Video Diffusion Models}} with {{An Expert Transformer}}.
\newblock In \emph{Int. Conf. Learn. Represent.}, 2025.

\bibitem[Yao et~al.(2025)Yao, Yang, and Wang]{yao2025reconstruction}
Jingfeng Yao, Bin Yang, and Xinggang Wang.
\newblock Reconstruction vs. generation: Taming optimization dilemma in latent diffusion models.
\newblock In \emph{Proceedings of the Computer Vision and Pattern Recognition Conference}, pp.\  15703--15712, 2025.

\bibitem[Yu et~al.(2024)Yu, Kwak, Jang, Jeong, Huang, Shin, and Xie]{yu2024representation}
Sihyun Yu, Sangkyung Kwak, Huiwon Jang, Jongheon Jeong, Jonathan Huang, Jinwoo Shin, and Saining Xie.
\newblock Representation alignment for generation: Training diffusion transformers is easier than you think.
\newblock \emph{arXiv preprint arXiv:2410.06940}, 2024.

\bibitem[Yuan et~al.(2025{\natexlab{a}})Yuan, Gao, Dai, Luo, Zhao, Zhang, Xie, Wei, Wang, Xiao, et~al.]{yuan2025native}
Jingyang Yuan, Huazuo Gao, Damai Dai, Junyu Luo, Liang Zhao, Zhengyan Zhang, Zhenda Xie, YX~Wei, Lean Wang, Zhiping Xiao, et~al.
\newblock Native sparse attention: Hardware-aligned and natively trainable sparse attention.
\newblock \emph{arXiv preprint arXiv:2502.11089}, 2025{\natexlab{a}}.

\bibitem[Yuan et~al.(2025{\natexlab{b}})Yuan, Wang, Sun, Zhang, and Lin]{yuan2025tarsier2}
Liping Yuan, Jiawei Wang, Haomiao Sun, Yuchen Zhang, and Yuan Lin.
\newblock Tarsier2: Advancing large vision-language models from detailed video description to comprehensive video understanding.
\newblock \emph{arXiv preprint arXiv:2501.07888}, 2025{\natexlab{b}}.

\bibitem[Zhai et~al.(2022)Zhai, Kolesnikov, Houlsby, and Beyer]{zhai2022scaling}
Xiaohua Zhai, Alexander Kolesnikov, Neil Houlsby, and Lucas Beyer.
\newblock Scaling vision transformers.
\newblock In \emph{Proceedings of the IEEE/CVF conference on computer vision and pattern recognition}, pp.\  12104--12113, 2022.

\bibitem[Zhang et~al.(2025{\natexlab{a}})Zhang, Li, Cheng, Hu, Yuan, Chen, Leng, Jiang, Zhang, Li, et~al.]{zhang2025videollama}
Boqiang Zhang, Kehan Li, Zesen Cheng, Zhiqiang Hu, Yuqian Yuan, Guanzheng Chen, Sicong Leng, Yuming Jiang, Hang Zhang, Xin Li, et~al.
\newblock Videollama 3: Frontier multimodal foundation models for image and video understanding.
\newblock \emph{arXiv preprint arXiv:2501.13106}, 2025{\natexlab{a}}.

\bibitem[Zhang et~al.(2025{\natexlab{b}})Zhang, Chen, Su, Ding, Stoica, Liu, and Zhang]{zhang2025fast}
Peiyuan Zhang, Yongqi Chen, Runlong Su, Hangliang Ding, Ion Stoica, Zhengzhong Liu, and Hao Zhang.
\newblock Fast video generation with sliding tile attention.
\newblock \emph{arXiv preprint arXiv:2502.04507}, 2025{\natexlab{b}}.

\bibitem[Zhang et~al.(2018)Zhang, Isola, Efros, Shechtman, and Wang]{zhang2018unreasonable}
Richard Zhang, Phillip Isola, Alexei~A Efros, Eli Shechtman, and Oliver Wang.
\newblock The unreasonable effectiveness of deep features as a perceptual metric.
\newblock In \emph{Proceedings of the IEEE conference on computer vision and pattern recognition}, pp.\  586--595, 2018.

\bibitem[Zhang et~al.(2025{\natexlab{c}})Zhang, Li, Chen, Ge, Sun, Zhang, Jiang, Yuan, Peng, and Luo]{zhang2025flashvideo}
Shilong Zhang, Wenbo Li, Shoufa Chen, Chongjian Ge, Peize Sun, Yida Zhang, Yi~Jiang, Zehuan Yuan, Binyue Peng, and Ping Luo.
\newblock Flashvideo: Flowing fidelity to detail for efficient high-resolution video generation.
\newblock \emph{arXiv preprint arXiv:2502.05179}, 2025{\natexlab{c}}.

\bibitem[Zhang et~al.(2025{\natexlab{d}})Zhang, Long, Qiu, Pan, Liu, Yao, and Mei]{zhang2025motionpro}
Zhongwei Zhang, Fuchen Long, Zhaofan Qiu, Yingwei Pan, Wu~Liu, Ting Yao, and Tao Mei.
\newblock Motionpro: A precise motion controller for image-to-video generation.
\newblock In \emph{Proceedings of the Computer Vision and Pattern Recognition Conference}, pp.\  27957--27967, 2025{\natexlab{d}}.

\bibitem[Zhao et~al.(2023)Zhao, Gu, Varma, Luo, Huang, Xu, Wright, Shojanazeri, Ott, Shleifer, et~al.]{zhao2023pytorch}
Yanli Zhao, Andrew Gu, Rohan Varma, Liang Luo, Chien-Chin Huang, Min Xu, Less Wright, Hamid Shojanazeri, Myle Ott, Sam Shleifer, et~al.
\newblock Pytorch fsdp: experiences on scaling fully sharded data parallel.
\newblock \emph{arXiv preprint arXiv:2304.11277}, 2023.

\end{thebibliography}
